\documentclass[10pt,twocolumn,letterpaper]{article}
\ifdefined\pdfminorversion\pdfminorversion=7\relax\fi

\usepackage{cvpr}

\usepackage{multirow}
\usepackage{colortbl}

\definecolor{cvprblue}{rgb}{0.21,0.49,0.74}
\definecolor{best}{RGB}{205,35,35}
\definecolor{second}{RGB}{35,75,195}
\definecolor{third}{RGB}{0,125,55}
\usepackage[pagebackref,breaklinks,colorlinks,allcolors=cvprblue]{hyperref}

\def\confName{CVPR}
\def\confYear{2027}

\title{TDD: Target--Discrepancy Distillation for Underwater Image Enhancement}

\author{
\textnormal{Zefan Feng$^{1}$ \quad Zhaocheng Yu$^{1}$ \quad Laibin Chang$^{2}$} \\
\textnormal{Kui Jiang$^{1}$ \quad Junjun Jiang$^{1}$ \quad Xiaopeng Fan$^{1}$} \\
\textnormal{$^{1}$Harbin Institute of Technology \quad $^{2}$Zhengzhou University} \\
\texttt{\{25s103429,yuzhaocheng\}@stu.hit.edu.cn} \\
\texttt{changlb@zzu.edu.cn} \\
\texttt{\{jiangkui,jiangjunjun,fxp\}@hit.edu.cn}
}

\begin{document}

\maketitle

\begin{abstract}
Underwater degradation changes image appearance and can compromise the semantic cues used to guide enhancement.
Identifying scene content alone does not explicitly specify the appearance to recover or the input-dependent changes associated with that target.
We propose \emph{target--discrepancy knowledge distillation (TDD)} for underwater image enhancement, which organizes guidance around these two aspects of the same restoration problem.
For each degraded--reference training pair, reference features from an underwater vision-language teacher provide target-appearance supervision, while paired feature residuals describe the observation's discrepancy relative to the same target. Both representations are jointly distilled into a lightweight guidance network that predicts them from the degraded image alone; neither the teacher nor reference images are required at inference.
To align their use with their supervisory roles, we introduce progressive knowledge fusion within a multi-scale enhancement network.
The estimated discrepancy first conditions spatial feature updates, after which target-appearance guidance refines the updated features and channel interactions consolidate their responses.
This correction-before-refinement design lets target-guided refinement build on discrepancy-conditioned features.
Experiments on six underwater benchmarks demonstrate competitive performance across full-reference and no-reference metrics.
Ablations and feature visualizations analyze the contributions of the knowledge source, joint distillation, and progressive fusion.

\end{abstract}

\section{Introduction}
\label{sec:introduction}

\begin{figure}[!t]
    \centering
    \includegraphics[width=\columnwidth]{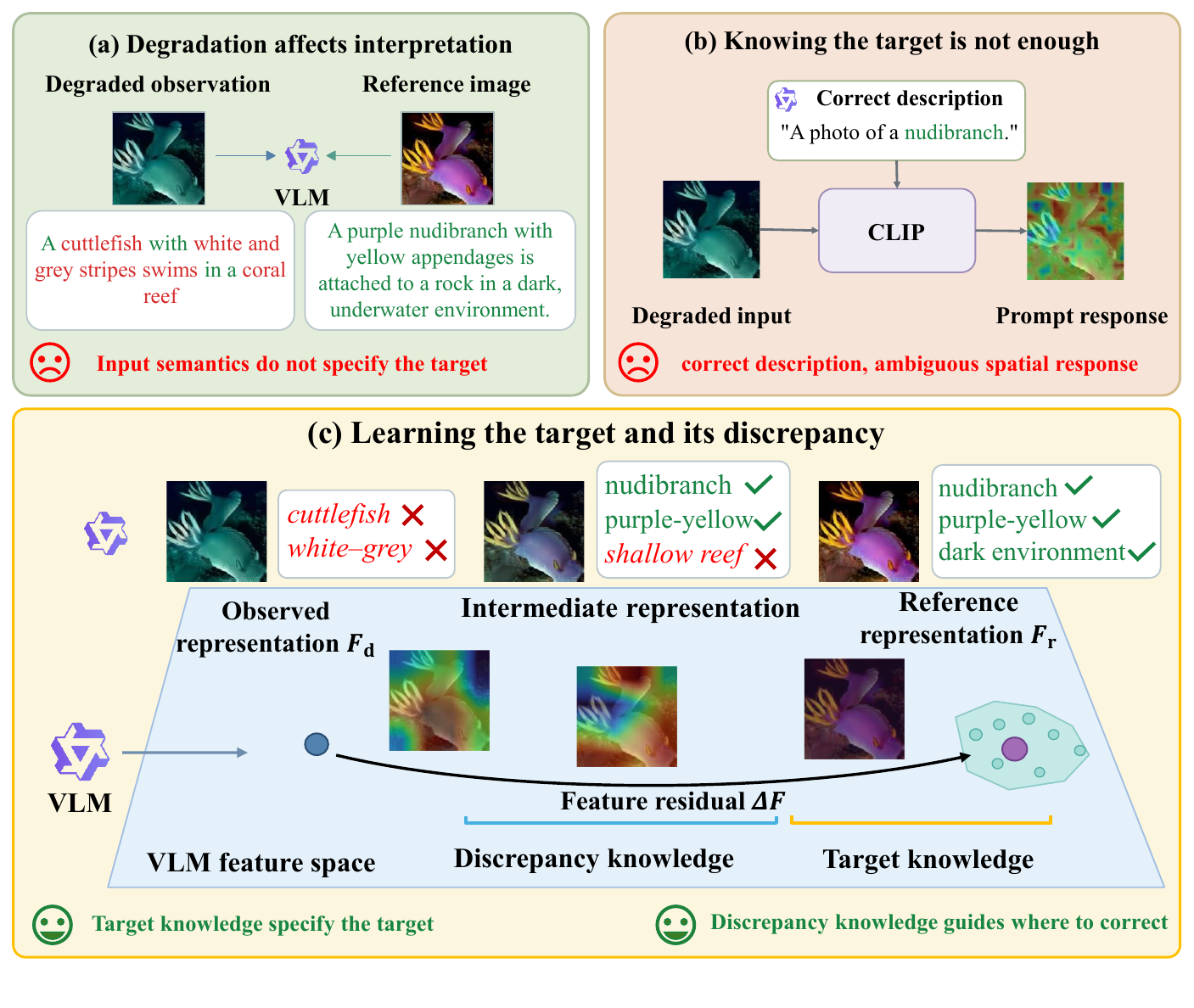}\vspace{-2mm}
    \caption{\textbf{Motivation for target--discrepancy guidance.}
    (a) The same object receives different semantic descriptions in a degraded image and its reference, illustrating the sensitivity of interpretation to appearance changes.
    (b) A correct object description can still yield an ambiguous spatial response.
    (c) Appearance variations and the corresponding feature-discrepancy responses motivate learning a reference-defined appearance target together with the observation's discrepancy from it.}\vspace{-3mm}
    \label{fig:motivation}
    
\end{figure}

Underwater image enhancement (UIE) aims to restore image appearance impaired by light absorption and scattering, which cause color distortion, reduced contrast, and loss of visible detail~\cite{50695,8578801}.
These effects also alter the color, texture, and contour cues used to interpret scene content.
Consequently, semantic guidance inferred from a degraded image can be affected by the very degradation it is intended to help correct.
As illustrated in Fig.~\ref{fig:motivation}(a), the same object receives different category and appearance descriptions in the degraded image and its reference, indicating the sensitivity of its semantic interpretation to appearance changes.
This sensitivity motivates guidance that explicitly connects the degraded observation to the appearance to be recovered.

Vision-language models (VLMs) have enabled increasingly content-aware UIE.
Semantic-sensitive enhancement~\cite{fan2026semanticsensitive} maps object descriptions to spatial guidance, LSFM-Diff~\cite{fan2026lsfmdiff} incorporates textual semantics into local feature fusion and attention, and DIE2UIE~\cite{wang2026die2uie} combines degradation-aware and semantic-aware prompts.
These methods establish useful connections between scene understanding and enhancement.
However, identifying an object does not uniquely specify its desired color, contrast, or local appearance.
Even a correct description can produce an ambiguous spatial response under degradation, as illustrated in Fig.~\ref{fig:motivation}(b).
The distinction is therefore between recognizing scene content and explicitly relating its observed appearance to a restoration target.

Feature-based guidance offers a more direct way to express this relation.
DiffRes~\cite{wang2025diffres} constructs restoration instructions from paired vision-language feature differences, while DATPRL-IR~\cite{dong2026datprlir} distills multimodal domain priors into learnable prompts.
More recently, ImIR~\cite{aslan2026imir} predicts clean-image instruction embeddings through a learned residual mapping from degraded embeddings.
Our focus is the joint learning and downstream use of these representations: we retain separate predictions of target appearance and observation discrepancy, and assign them distinct roles in enhancement.
This connects reference-based supervision with an ordered process of discrepancy-conditioned feature correction and target-guided refinement.

To this end, we propose \emph{target--discrepancy knowledge distillation (TDD)}.
We construct supervision in a shared feature space of NAUTILUS~\cite{xu2025nautilus}, a multimodal model developed for underwater scene understanding.
For each degraded--reference training pair, the reference features define \emph{target-appearance knowledge}, while the paired feature residuals define \emph{discrepancy knowledge} relative to that same target.
The two representations provide different supervisory views of a common restoration problem: the appearance to approach and the observation-dependent deviation from it.
A lightweight guidance network jointly learns to predict both representations from the degraded image.
This transfers reference-based supervision into input-conditioned guidance, so neither reference images nor the teacher are required at inference. 

The intended roles of these representations also inform how they are integrated into the enhancement network.
We introduce a \emph{progressive knowledge fusion} module that first combines the estimated discrepancy with spatial image features and local visual cues to condition feature updates.
Target-appearance knowledge then guides refinement of the updated features, followed by channel interactions that consolidate the resulting responses.
This ordering encodes a \emph{correction-before-refinement} inductive bias: appearance refinement operates on features already conditioned on the estimated input-to-target discrepancy.
Applying the module across multiple scales brings this division of roles to both global appearance recovery and local detail reconstruction.

Our contributions are summarized as follows:
\begin{itemize}
    \item We formulate target--discrepancy knowledge distillation for underwater image enhancement, jointly transferring reference features and paired feature residuals into an input-conditioned guidance network. This formulation associates an appearance target with the discrepancy of its corresponding observation.
    \item We introduce progressive knowledge fusion that uses discrepancy predictions to condition spatial feature updates before target-guided refinement, translating the two supervisory roles into an ordered multi-scale enhancement process.

    \item We conduct quantitative evaluation on six underwater benchmarks using full-reference and no-reference metrics, with additional qualitative comparisons. Ablations examine guidance representations, image--knowledge correspondence, and the contributions and ordering of spatial knowledge fusion.
\end{itemize}

\section{Related Work}
\label{sec:related-work}

\subsection{Underwater Image Enhancement}

Underwater image enhancement addresses color distortion,
contrast reduction, and detail loss caused by wavelength-dependent
attenuation and scattering~\cite{8578801}. Conventional approaches exploit
physical priors and wavelength compensation~\cite{6104148,7574330,8954437},
or combine color correction with contrast enhancement~\cite{ancuti2012enhancing,9788535}.
Learning-based methods explore multi-color-space representations~\cite{li2021underwater,wang2021uiec},
multi-scale feature modeling~\cite{peng2023u,11330175},
prior-guided adaptation~\cite{mu2023generalized,11267029}, and
semi-supervised or contrastive learning~\cite{huang2023contrastive,zhou2024hclr}.
Other studies investigate diffusion and frequency-domain processing for
color and detail recovery~\cite{tang2023underwater,zhao2024wavelet,11353942},
as well as enhancement for downstream visual perception~\cite{9001231,11433506}.
These methods improve visual recovery, but jointly representing
target appearance and its corresponding observation discrepancy
as explicit guidance remains less explored. This is particularly
relevant underwater, where degradation can also distort semantic
cues. Our method addresses this issue through reference-derived
target--discrepancy guidance.

\begin{figure*}[!htp]
    \centering
    \includegraphics[width=1.0\textwidth]{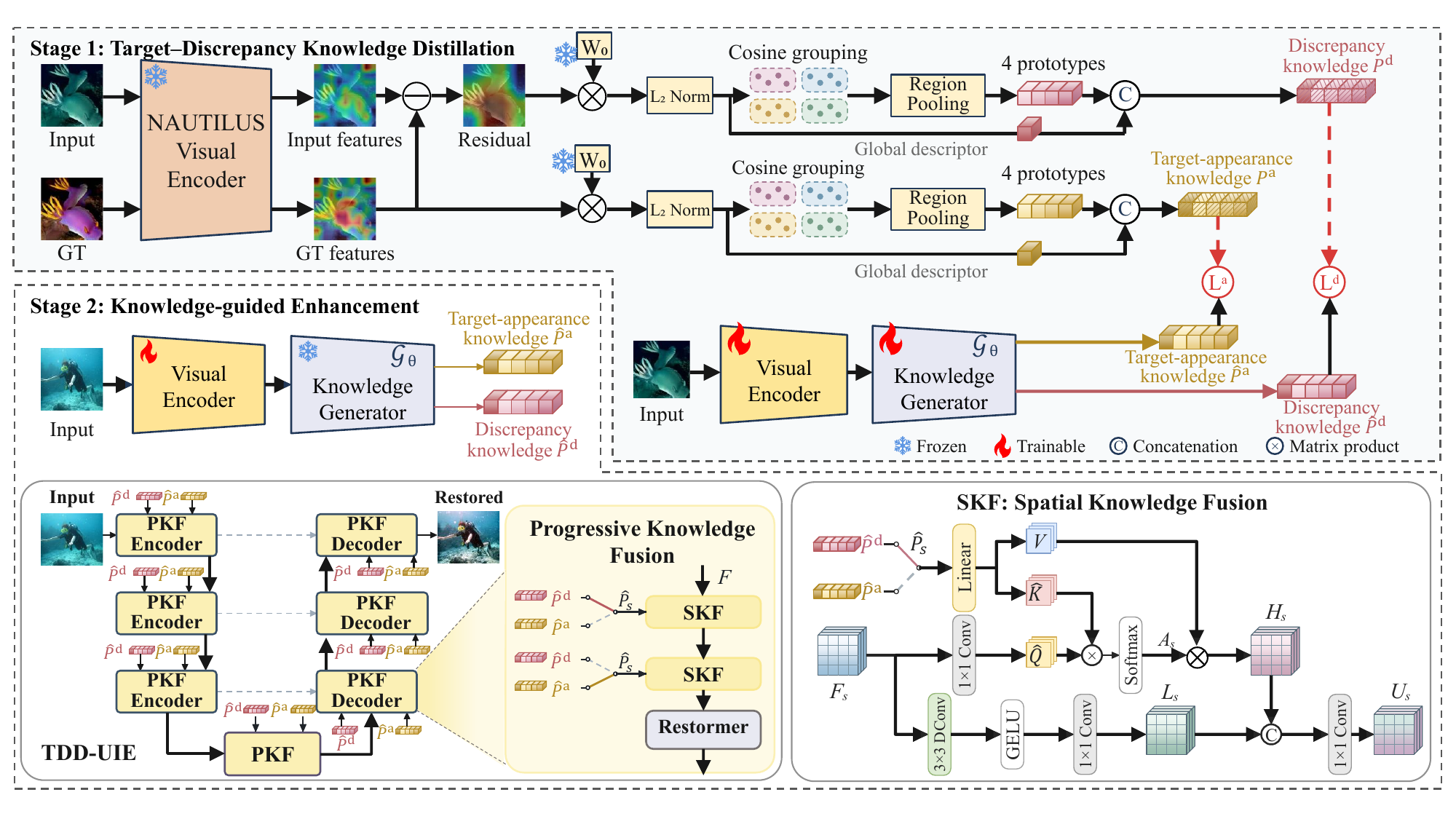}\vspace{-2mm}
    \caption{Overview of TDD. Stage 1 distills teacher-derived target-appearance knowledge $P^{\mathrm a}$ and discrepancy knowledge $P^{\mathrm d}$ into predictions from the degraded input. Stage 2 freezes the knowledge generator and optimizes the visual encoder and restoration network with image restoration supervision. Each PKF module sequentially performs discrepancy-conditioned SKF, target-guided SKF, and Restormer-based feature consolidation.}\vspace{-4mm}
    
    \label{fig:pipeline}
\end{figure*}

\subsection{Vision-Language-Guided Restoration}
Prompt-based methods condition restoration on task or content
information. PromptIR~\cite{potlapalli2023promptir} learns prompts for degradation adaptation,
while InstructIR~\cite{conde2024instructir} follows natural-language instructions.
VLM-based methods further provide content and degradation cues.
DA-CLIP~\cite{luo2024daclip} derives complementary content and degradation embeddings
using a controller with a frozen CLIP encoder. For UIE,
semantic-sensitive enhancement~\cite{fan2026semanticsensitive} constructs spatial guidance from
text--image similarities, LSFM-Diff~\cite{fan2026lsfmdiff} modulates features with
semantic information, and DIE2UIE~\cite{wang2026die2uie} combines degradation-aware
prompts with object- and scene-level semantics.

Feature-based approaches offer another route. DiffRes~\cite{wang2025diffres} derives
restoration instructions from paired visual feature differences,
whereas DATPRL-IR~\cite{dong2026datprlir} transfers CLIP representations of high-quality
image descriptions into learnable domain prompts. NAUTILUS~\cite{xu2025nautilus}
provides underwater multimodal representations through instruction
tuning at image, region, and object levels. TDD jointly distills
reference features and paired feature discrepancies within this
teacher space, then applies discrepancy-conditioned fusion before
target-guided refinement.

\section{Method}
\label{sec:method}

\subsection{Overview}
\label{sec:overview}

As illustrated in Fig.~\ref{fig:pipeline}, TDD integrates target--discrepancy knowledge distillation with Progressive Knowledge Fusion (PKF) in a two-stage training framework. In Stage 1, a frozen underwater vision-language teacher extracts paired representations from degraded and reference images to construct target-appearance and discrepancy knowledge. A lightweight guidance network, comprising a ResNet-18 visual encoder $E_\phi$ and a knowledge generator $\mathcal G_\theta$, learns to predict both representations from the degraded image. In Stage 2, $\mathcal G_\theta$ is frozen, while $E_\phi$ and a U-shaped restoration network are jointly optimized using image restoration supervision. The predicted knowledge guides multi-scale restoration through PKF, which first performs discrepancy-conditioned feature correction and then target-guided appearance refinement. At inference, the framework operates directly on the degraded image, without requiring reference images or the pretrained teacher.

\subsection{Target--Discrepancy Knowledge Distillation}
\label{sec:knowledge-distillation}

\textbf{Knowledge Construction.}
Given a degraded image $I$ and its corresponding reference $Y$, we extract the visual tokens $Z_I$ and $Z_Y$ using the frozen NAUTILUS model~\cite{xu2025nautilus}, followed by feature-wise $\ell_2$ normalization. We construct two complementary forms of knowledge in the same teacher feature space:
\begin{equation}
P_{\mathrm{raw}}^{\mathrm a}=Z_Y,
\qquad
P_{\mathrm{raw}}^{\mathrm d}=Z_I-Z_Y.
\label{eq:raw-knowledge}
\end{equation}
Here, $P_{\mathrm{raw}}^{\mathrm a}$ encodes the reference-defined target appearance, whereas $P_{\mathrm{raw}}^{\mathrm d}$ characterizes the degraded observation's feature discrepancy from that target. Sharing the same reference anchor links the desired appearance with its corresponding observation-dependent deviation.

\noindent \textbf{Compact Knowledge Representation.}
To retain regional variations while reducing the cost of distillation and spatial association, we represent each knowledge branch using a global descriptor and $K$ regional prototypes. Both branches undergo a shared fixed projection $W_0$ and $\ell_2$ normalization. The projected regional features are grouped according to cosine similarity and aggregated as
\begin{equation}
p_k^r=\operatorname{Pool}(\Omega_k^r),
\qquad r\in\{\mathrm d,\mathrm a\},
\label{eq:regional-prototype}
\end{equation}
where $\Omega_k^r$ denotes the $k$-th group in branch $r$. Within each group, we pool the regional features by weighted averaging followed by $\ell_2$ normalization. We use uniform weights for target-appearance features and residual-magnitude-based weights for discrepancy features, emphasizing regions with larger feature deviations. The resulting compact representation is
\begin{equation}
P^r=[\,p_{\mathrm g}^r;p_1^r;\ldots;p_K^r\,]
\in\mathbb R^{(K+1)\times d},
\label{eq:knowledge-target}
\end{equation}
where $d$ is the projected feature dimension. The global descriptors $p_{\mathrm g}^{\mathrm a}$ and $p_{\mathrm g}^{\mathrm d}$ are obtained by applying the same projection and normalization to the global reference feature and paired-feature difference, respectively. We set $K=4$ in all experiments.

\noindent \textbf{Joint Knowledge Distillation.}
In Stage 1, the guidance network predicts both knowledge representations from the degraded input:
$(\widehat P^{\mathrm d},\widehat P^{\mathrm a})
=
\mathcal G_\theta(E_\phi(I)),$
where $E_\phi$ denotes the ResNet-18 visual encoder and $\mathcal G_\theta$ denotes the knowledge generator. The predicted representations have the same structures and dimensions as their corresponding teacher-derived targets. We jointly supervise the two branches using a cosine-distance objective:
\begin{equation}
\begin{aligned}
\mathcal L^r
&=\mathcal D_{\cos}(\widehat P^r,P^r),
\qquad r\in\{\mathrm d,\mathrm a\},\\
\mathcal L_{\mathrm{distill}}
&=\mathcal L^{\mathrm d}+\mathcal L^{\mathrm a},
\end{aligned}
\label{eq:distillation}
\end{equation}
where $\mathcal D_{\cos}$ computes the weighted mean cosine distance between corresponding knowledge vectors. Global descriptors are aligned directly, whereas regional prototypes are compared after similarity-based matching. The two objectives provide complementary supervision: target-appearance distillation transfers reference-derived appearance representations, while discrepancy distillation teaches the network to estimate the corresponding observation-to-target feature deviations.

\noindent \textbf{Transfer to Enhancement.}
After Stage 1, we freeze $\mathcal G_\theta$ and train $E_\phi$ jointly with the restoration network using image restoration supervision. Although the generator parameters remain fixed, its predictions remain input-adaptive through the trainable visual encoder. The resulting knowledge representations condition the restoration process through PKF.

\subsection{Progressive Knowledge Fusion}
\label{sec:progressive-fusion}
Target-appearance and discrepancy knowledge encode complementary aspects of restoration and therefore play distinct roles. Discrepancy knowledge characterizes the deviation from the desired appearance and guides feature correction, whereas target-appearance knowledge specifies the restoration objective and guides subsequent refinement. We implement this ordered interaction through PKF.

Given image features $F$, PKF sequentially performs discrepancy-conditioned and target-guided Spatial Knowledge Fusion (SKF), followed by feature consolidation through a Restormer block $\mathcal R$~\cite{Zamir2022Restormer}:
\begin{equation}
\begin{aligned}
F_{\mathrm d}
&=\operatorname{SKF}_{\mathrm d}
(F,\widehat P^{\mathrm d}),~~
F_{\mathrm a}
=\operatorname{SKF}_{\mathrm a}
(F_{\mathrm d},\widehat P^{\mathrm a}),\\
F_{\mathrm{out}}
&=\mathcal R(F_{\mathrm a}).
\end{aligned}
\label{eq:progressive-fusion}
\end{equation}
Here, $F_{\mathrm d}$ denotes discrepancy-conditioned features and $F_{\mathrm a}$ denotes their target-guided refinement. Importantly, the second SKF operates on $F_{\mathrm d}$ rather than the original features, enforcing a correction-before-refinement order. PKF is deployed across the encoder, bottleneck, and decoder to provide complementary guidance at multiple spatial scales.

\begin{table*}[!t]
\centering
\captionsetup{skip=10pt}
\caption{Quantitative comparisons on three full-reference test sets. The top three distinct values are marked in \textcolor{best}{red}, \textcolor{second}{blue}, and \textcolor{third}{green}.}\vspace{-2mm}
\label{tab:paired}
\footnotesize
\setlength{\tabcolsep}{1.4pt}
\renewcommand{\arraystretch}{1.0}
\begin{tabular*}{0.90\textwidth}{@{\extracolsep{\fill}}lc*{12}{c}@{}}
\toprule
\multirow{2}{*}{Method} & \multirow{2}{*}{Venue}
 & \multicolumn{3}{c}{\textbf{UIEB}}
 & \multicolumn{3}{c}{\textbf{LSUI}}
 & \multicolumn{3}{c}{\textbf{EUVP}}
 & \multicolumn{3}{c}{\textbf{Efficiency}} \\
\cmidrule(lr){3-5}\cmidrule(lr){6-8}\cmidrule(lr){9-11}\cmidrule(lr){12-14}
 & & \scriptsize PSNR $\uparrow$ & \scriptsize SSIM $\uparrow$ & \scriptsize UCIQE $\uparrow$
 & \scriptsize PSNR $\uparrow$ & \scriptsize SSIM $\uparrow$ & \scriptsize UCIQE $\uparrow$
 & \scriptsize PSNR $\uparrow$ & \scriptsize SSIM $\uparrow$ & \scriptsize UCIQE $\uparrow$
 & \scriptsize Params (M) $\downarrow$ & \scriptsize FLOPs (G) $\downarrow$ & \scriptsize Time (s) $\downarrow$ \\
\midrule
U-Shape~\cite{peng2023u} & TIP'23
 & 20.54 & 0.772 & 0.575
 & 24.16 & 0.846 & 0.575
 & 25.48 & 0.830 & 0.580
 & 65.60 & 66.20 & 0.063 \\
Semi-UIR~\cite{huang2023contrastive} & CVPR'23
 & 23.49 & 0.860 & 0.612
 & 27.46 & 0.878 & 0.590
 & 26.85 & 0.876 & 0.586
 & \textcolor{best}{3.31} & 72.88 & 0.785 \\
CECF~\cite{cong2024underwater} & AAAI'24
 & 21.59 & 0.815 & 0.609
 & 23.79 & 0.839 & 0.592
 & 23.36 & 0.813 & \textcolor{third}{0.592}
 & 39.69 & 83.58 & \textcolor{best}{0.003} \\
WF-Diff~\cite{zhao2024wavelet} & CVPR'24
 & 22.78 & 0.912 & \textcolor{best}{0.619}
 & 27.49 & \textcolor{third}{0.894} & 0.592
 & 27.01 & 0.876 & 0.585
 & 18.48 & 570.09 & 0.146 \\
HCLR~\cite{zhou2024hclr} & IJCV'24
 & 21.98 & 0.887 & 0.598
 & 28.74 & 0.889 & 0.589
 & 27.86 & 0.864 & 0.583
 & \textcolor{third}{4.87} & 401.90 & \textcolor{third}{0.054} \\
MaIR~\cite{li2025mair} & CVPR'25
 & \textcolor{second}{25.11} & \textcolor{second}{0.923} & \textcolor{second}{0.618}
 & \textcolor{second}{29.76} & \textcolor{best}{0.908} & \textcolor{best}{0.597}
 & \textcolor{second}{29.90} & \textcolor{second}{0.899} & \textcolor{best}{0.594}
 & 7.78 & \textcolor{best}{12.59} & 0.377 \\
EVSSM~\cite{kong2025efficient} & CVPR'25
 & \textcolor{third}{24.93} & \textcolor{third}{0.916} & \textcolor{third}{0.617}
 & \textcolor{third}{29.49} & \textcolor{second}{0.902} & \textcolor{second}{0.594}
 & \textcolor{third}{29.67} & \textcolor{third}{0.885} & 0.588
 & 17.10 & 126.00 & 0.279 \\
IA2U~\cite{11267029} & TIP'25
 & 22.53 & 0.888 & 0.605
 & 26.13 & 0.883 & \textcolor{third}{0.593}
 & 26.29 & 0.870 & 0.589
 & 21.70 & \textcolor{third}{33.32} & 0.081 \\
PyUIE~\cite{11330175} & TMM'26
 & 22.94 & 0.898 & 0.604
 & 26.92 & 0.882 & 0.590
 & 27.07 & 0.870 & 0.585
 & \textcolor{second}{4.45} & \textcolor{second}{12.92} & \textcolor{second}{0.039} \\
DiffColor~\cite{11353942} & TMM'26
 & 23.36 & 0.902 & 0.611
 & 28.07 & 0.891 & 0.592
 & 28.58 & 0.884 & 0.586
 & 22.16 & 87.95 & 0.157 \\
DTIUIE~\cite{11433506} & TIP'26
 & 24.57 & 0.911 & 0.608
 & 28.18 & 0.893 & 0.584
 & 28.10 & 0.880 & 0.569
 & 50.81 & 385.93 & 0.112 \\
\midrule
\textbf{TDD (Ours)} & --
 & \textcolor{best}{26.16} & \textcolor{best}{0.928} & \textcolor{second}{0.618}
 & \textcolor{best}{29.81} & \textcolor{best}{0.908} & \textcolor{best}{0.597}
 & \textcolor{best}{30.46} & \textcolor{best}{0.901} & \textcolor{second}{0.593}
 & 41.86 & 185.74 & 0.093 \\
\bottomrule
\end{tabular*}\vspace{-2mm}
\end{table*}

\begin{figure*}[!htbp]
\centering
\includegraphics[width=0.99\textwidth]{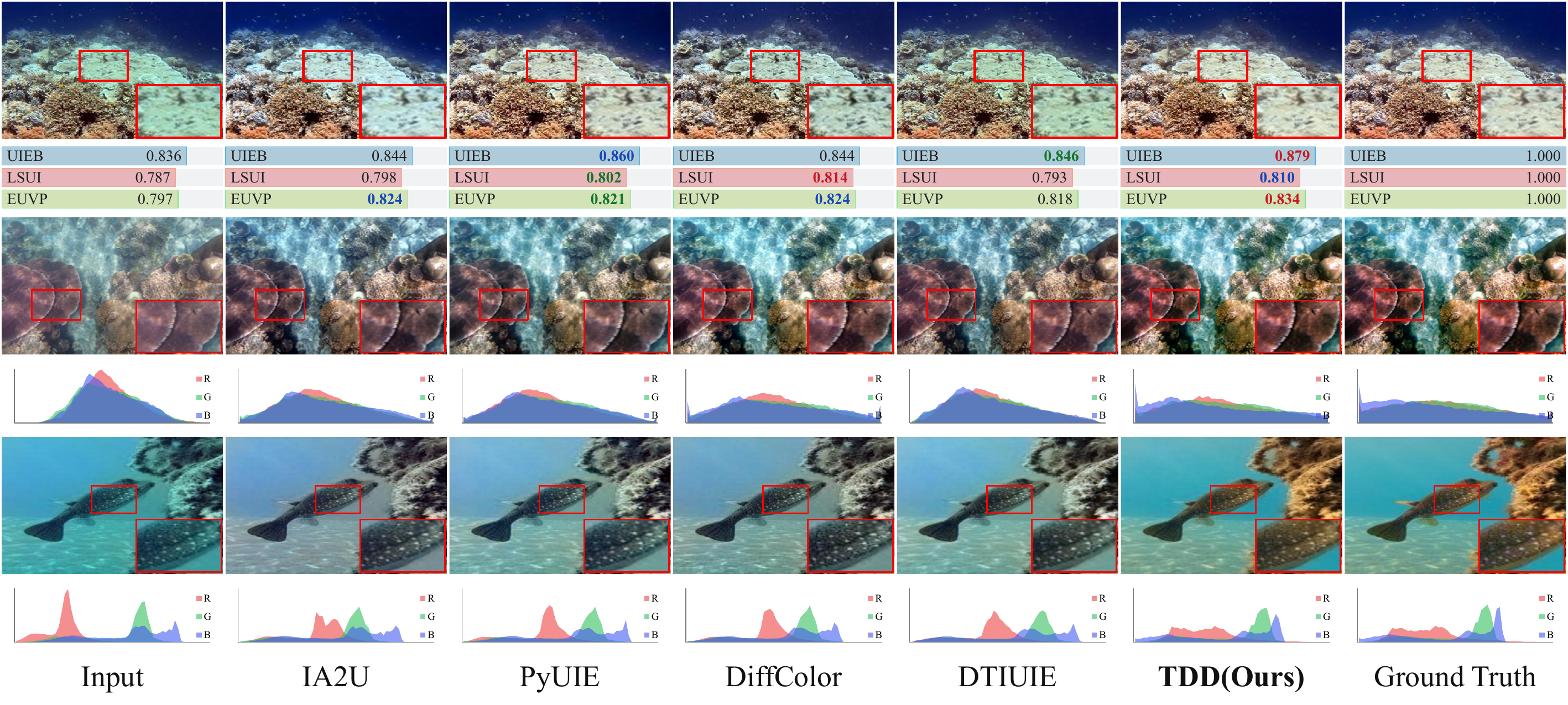}\vspace{-2mm}
    \caption{Visual comparisons with local enlargements and color histograms. The
    bars report the mean semantic similarity between generated captions and reference-image captions on UIEB, LSUI, and EUVP. The top three distinct scores at the displayed precision are highlighted in \textcolor{best}{red}, \textcolor{second}{blue}, and \textcolor{third}{green}.}\vspace{-4mm}
    \label{fig:paired-visual-comparison}
\end{figure*}

\noindent \textbf{Spatial Knowledge Association.}
Each SKF step establishes spatial correspondences between image features and the selected knowledge representation. Let $s\in\{1,2\}$ index the two steps, with $\widehat P_1=\widehat P^{\mathrm d}$ and $\widehat P_2=\widehat P^{\mathrm a}$. After normalization, channel mapping, and spatial flattening, the current image features are represented as $F_s\in\mathbb R^{N_s\times c_s}$, where $N_s$ and $c_s$ denote the number of spatial locations and the internal channel dimension, respectively. Image features serve as queries, while knowledge vectors provide keys and values. 
The spatial association is computed as
\begin{equation}
A_s=
\operatorname{Softmax}
\left(
\mathcal S_s
(F_sW_Q^s,\widehat P_sW_K^s)
\right),
\label{eq:spatial-association}
\end{equation}
followed by knowledge aggregation:
$H_s=A_s(\widehat P_sW_V^s),$
where $W_Q^s$, $W_K^s$, and $W_V^s$ are learnable projection matrices. Query projection uses a $1\times1$ convolution, while key and value projections use linear layers. The similarity function $\mathcal S_s$ computes cosine similarity calibrated by a learnable scale and predicted knowledge weights. Softmax is applied over the $K+1$ knowledge vectors, yielding $A_s\in\mathbb R^{N_s\times(K+1)}$ and spatial guidance $H_s\in\mathbb R^{N_s\times c_s}$. Thus, each image location adaptively retrieves relevant knowledge according to its current feature representation. Separate projection parameters are used for the two SKF steps to preserve their distinct association patterns.

\noindent \textbf{Local Feature Fusion.}
To complement knowledge-derived guidance with local structural and textural evidence, we extract local features $L_s$ through a $3\times3$ depth-wise convolution, GELU activation, and a $1\times1$ convolution. These features are concatenated with spatial guidance and projected to the internal channel dimension:
$U_s=
\operatorname{Conv}_{1\times1}
\left(
[L_s;H_s]
\right),$
where $U_s$ denotes the fused representation. This operation combines spatially associated knowledge with local visual evidence, enabling the two SKF steps to perform complementary feature updates. Specifically, discrepancy-guided fusion conditions spatial feature correction, whereas target-guided fusion refines the corrected representation toward the desired appearance. The subsequent Restormer block further consolidates the fused responses through feature interactions. This sequential design preserves the complementary roles of the two knowledge representations throughout multi-scale restoration.Further details on knowledge construction and the training procedure are provided in the supplementary material.

\section{Experiments}
\label{sec:experiments}

\subsection{Experimental Setup}

\noindent \textbf{Datasets and Metrics.}
We train TDD on 4,302 UIEB--LSUI image pairs with cached knowledge targets~\cite{li2019underwater,peng2023u}. Test sets disjoint from training contain 90 UIEB, 400 LSUI, and 481 EUVP images. No-reference quantitative evaluation uses UCCS, Sea-thru, and C60; additional qualitative fusion comparisons include U45. On the paired test sets, we measure reconstruction fidelity with PSNR and SSIM~\cite{wang2004image} and assess the enhanced images with UCIQE~\cite{yang2015underwater}, a no-reference metric. On the unpaired test sets, we report UCIQE, UIQM~\cite{panetta2015human}, NIMA~\cite{8352823}, and CLIP-IQA~\cite{wang2023clipiqa}. We additionally report the mean semantic similarity between captions generated from enhanced images and those generated from their corresponding reference images on UIEB, LSUI, and EUVP.

\noindent \textbf{Implementation Details.}
We train in PyTorch on RTX 3080 GPUs with global batch size 14 and $256\times256$ images, augmented by random rotations in multiples of $90^\circ$ and horizontal flips. The backbone is trained from scratch; the visual encoder and knowledge generator are initialized with the weights learned during Stage 1 distillation, with the generator frozen in Stage 2. AdamW uses initial learning rates of $10^{-4}$ for the backbone and $10^{-5}$ for the encoder, with a 20-epoch warm-up followed by cosine decay over 600 epochs in total. We set the base width to 64, $K=4$, and seed to 100.

\begin{table*}[!t]
\centering
\captionsetup{skip=10pt}
\caption{Quantitative comparisons on three no-reference test sets. The top three results are marked in \textcolor{best}{red}, \textcolor{second}{blue}, and \textcolor{third}{green}, respectively.}\vspace{-2mm}
\label{tab:no-ref}
\footnotesize
\setlength{\tabcolsep}{1.4pt}
\renewcommand{\arraystretch}{1.0}
\begin{tabular*}{0.90\textwidth}{@{\extracolsep{\fill}}lc*{12}{c}@{}}
\toprule
\multirow{2}{*}{Method} & \multirow{2}{*}{Venue} & \multicolumn{4}{c}{\textbf{UCCS}} & \multicolumn{4}{c}{\textbf{Sea-thru}} & \multicolumn{4}{c}{\textbf{C60}} \\
\cmidrule(lr){3-6}\cmidrule(lr){7-10}\cmidrule(lr){11-14}
 & & \scriptsize UCIQE $\uparrow$ & \scriptsize UIQM $\uparrow$ & \scriptsize NIMA $\uparrow$ & \scriptsize CLIP-IQA $\uparrow$
 & \scriptsize UCIQE $\uparrow$ & \scriptsize UIQM $\uparrow$ & \scriptsize NIMA $\uparrow$ & \scriptsize CLIP-IQA $\uparrow$
 & \scriptsize UCIQE $\uparrow$ & \scriptsize UIQM $\uparrow$ & \scriptsize NIMA $\uparrow$ & \scriptsize CLIP-IQA $\uparrow$ \\
\midrule
U-Shape~\cite{peng2023u} & TIP'23 & 0.547 & \textcolor{second}{3.044} & 3.411 & 0.276 & 0.466 & 2.835 & 3.386 & 0.302 & 0.545 & 2.660 & 4.011 & 0.457 \\
Semi-UIR~\cite{huang2023contrastive} & CVPR'23 & 0.571 & 2.951 & 3.480 & 0.318 & 0.507 & 2.919 & 4.263 & 0.444 & \textcolor{third}{0.585} & 2.648 & 4.304 & 0.556 \\
CECF~\cite{cong2024underwater} & AAAI'24 & 0.568 & 2.948 & 3.436 & 0.279 & \textcolor{second}{0.536} & \textcolor{third}{2.967} & 3.770 & 0.417 & 0.583 & \textcolor{third}{2.695} & 3.923 & 0.515 \\
WF-Diff~\cite{zhao2024wavelet} & CVPR'24 & \textcolor{best}{0.576} & \textcolor{third}{3.041} & \textcolor{second}{3.610} & 0.325 & 0.501 & \textcolor{second}{3.028} & 4.391 & \textcolor{second}{0.598} & 0.584 & \textcolor{second}{2.715} & \textcolor{second}{4.346} & 0.538 \\
HCLR~\cite{zhou2024hclr} & IJCV'24 & 0.565 & 2.996 & 3.415 & 0.296 & 0.466 & 2.141 & 3.861 & 0.547 & 0.566 & \textcolor{best}{2.737} & 4.138 & \textcolor{second}{0.573} \\
MaIR~\cite{li2025mair} & CVPR'25 & 0.555 & 2.926 & \textcolor{best}{3.657} & 0.356 & \textcolor{third}{0.528} & 2.217 & 4.191 & \textcolor{third}{0.592} & \textcolor{best}{0.589} & 2.564 & \textcolor{best}{4.354} & 0.562 \\
EVSSM~\cite{kong2025efficient} & CVPR'25 & \textcolor{third}{0.572} & 2.963 & 3.565 & 0.311 & 0.485 & 2.823 & \textcolor{third}{4.403} & 0.572 & 0.584 & 2.640 & 4.302 & \textcolor{third}{0.564} \\
IA2U~\cite{11267029} & TIP'25 & 0.564 & 3.005 & 3.424 & \textcolor{third}{0.409} & 0.519 & 2.764 & \textcolor{second}{4.458} & 0.541 & 0.577 & 2.689 & 4.129 & 0.534 \\
PyUIE~\cite{11330175} & TMM'26 & 0.562 & 2.959 & 3.412 & 0.352 & 0.467 & 2.324 & 3.963 & 0.354 & 0.575 & 2.543 & 4.054 & 0.516 \\
DiffColor~\cite{11353942} & TMM'26 & 0.569 & 2.955 & 3.448 & 0.297 & 0.471 & 2.851 & 3.826 & 0.534 & 0.573 & 2.547 & 4.003 & 0.504 \\
DTIUIE~\cite{11433506} & TIP'26 & 0.564 & \textcolor{best}{3.071} & 3.462 & \textcolor{second}{0.420} & 0.496 & 2.892 & 3.901 & 0.518 & 0.572 & 2.651 & 4.127 & 0.533 \\
\midrule
\textbf{TDD (Ours)} & -- & \textcolor{second}{0.573} & 2.995 & \textcolor{third}{3.606} & \textcolor{best}{0.444} & \textcolor{best}{0.538} & \textcolor{best}{3.114} & \textcolor{best}{4.600} & \textcolor{best}{0.713} & \textcolor{second}{0.586} & 2.659 & \textcolor{third}{4.327} & \textcolor{best}{0.591} \\
\bottomrule
\end{tabular*}\vspace{0pt}
\end{table*}

\begin{figure*}[!t]
\centering
\captionsetup{skip=10pt}
\includegraphics[width=\textwidth]{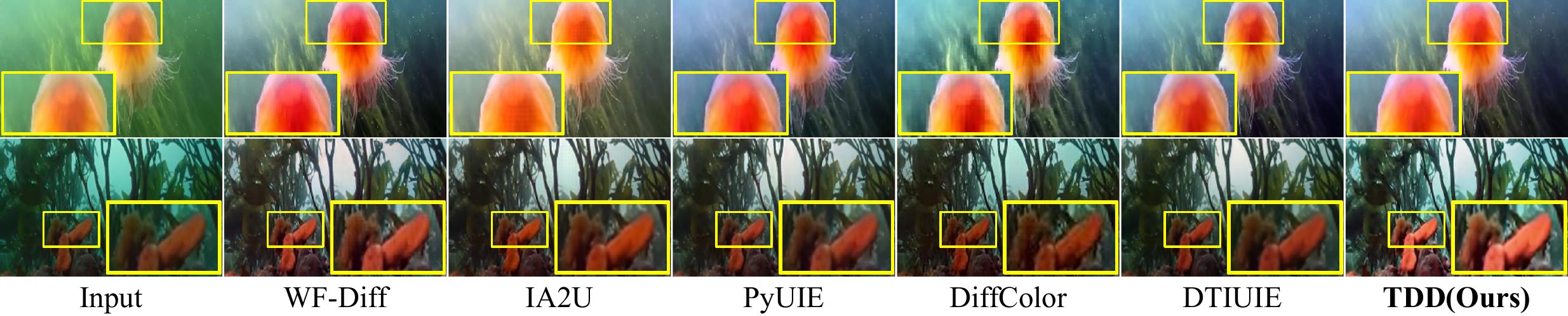}\vspace{-2mm}
    \caption{Visual comparisons on the C60 dataset with local enlargements.}\vspace{-4mm}
    \label{fig:c60-visual-comparison}
\end{figure*}

\subsection{Comparison with State-of-the-Art Methods}
\label{sec:comparison}
\noindent \textbf{Compared Methods.}
We compare TDD with 11 methods for underwater image enhancement and image restoration:
U-Shape~\cite{peng2023u}, Semi-UIR~\cite{huang2023contrastive}, CECF~\cite{cong2024underwater}, WF-Diff~\cite{zhao2024wavelet}, HCLR~\cite{zhou2024hclr}, MaIR~\cite{li2025mair}, EVSSM~\cite{kong2025efficient}, IA2U~\cite{11267029}, PyUIE~\cite{11330175}, DiffColor~\cite{11353942}, and DTIUIE~\cite{11433506}. These baselines cover diverse feature modeling and enhancement strategies.

\noindent \textbf{Full-reference Comparison.}
As shown in Table~\ref{tab:paired}, TDD achieves the highest PSNR on UIEB, LSUI, and EUVP, outperforming the next-best method by 1.05, 0.05, and 0.56~dB, respectively. It achieves the highest SSIM on UIEB and EUVP and ties with MaIR on LSUI at the reported precision. In UCIQE, TDD ties with MaIR for second on UIEB (0.618) and for first on LSUI (0.597), and ranks second on EUVP (0.593). The leading PSNR and SSIM on EUVP support generalization to a benchmark outside the joint training set.

\noindent \textbf{No-reference Comparison.}
Table~\ref{tab:no-ref} further evaluates enhancement quality on UCCS, Sea-thru, and C60. TDD ranks first in six of the twelve dataset--metric combinations, including CLIP-IQA on all three datasets and all four metrics on Sea-thru. It also ranks second in UCIQE and third in NIMA on both UCCS and C60, demonstrating competitive performance across multiple quality criteria.

\noindent \textbf{Visual Comparison.}
In Fig.~\ref{fig:paired-visual-comparison}, TDD recovers warmer fish and rock tones closer to the reference while retaining the blue--cyan appearance of the surrounding water; several competing results remain comparatively gray. The RGB histograms show closer agreement with the reference color distributions, providing complementary evidence of overall color recovery. For caption-based semantic similarity, TDD achieves the highest scores of 0.879 on UIEB and 0.834 on EUVP.

Figure~\ref{fig:c60-visual-comparison} shows that TDD reduces the green veil on C60 while preserving fine jellyfish tentacles and foreground structures in the kelp scene.

\subsection{Ablation Study and Analysis}
\label{sec:ablation}

\noindent\textbf{Guidance Representation.} We test three text-based alternatives to TDD using the same enhancement architecture. \emph{Text prompts} use CLIP to encode predefined descriptions of underwater objects and match them to image features by cosine similarity. \emph{Dual text prompts} also describe common degradations. For \emph{Fine-tuned CLIP}, an underwater multimodal large language model generates image-specific descriptions for both fine-tuning and test-time prompting.

\begin{table}[!tb]
\centering
\captionsetup{skip=10pt}
\caption{Guidance representations on UIEB and LSUI. Best in \textcolor{best}{red}.}\vspace{-2mm}
\label{tab:guidance-ablation-review}
\footnotesize
\setlength{\tabcolsep}{3pt}
\renewcommand{\arraystretch}{0.90}
\begin{tabular*}{0.85\columnwidth}{@{\extracolsep{\fill}}lcccc@{}}
    \toprule
    \multirow{2}{*}{Guidance}
    & \multicolumn{2}{c}{UIEB} & \multicolumn{2}{c}{LSUI} \\
    \cmidrule(lr){2-3}\cmidrule(lr){4-5}
    & PSNR $\uparrow$ & SSIM $\uparrow$ & PSNR $\uparrow$ & SSIM $\uparrow$ \\
    \midrule
    Text prompts & 24.73 & 0.917 & 29.17 & 0.904 \\
    Dual text prompts & 25.26 & 0.922 & 29.20 & 0.901 \\
    Fine-tuned CLIP & 25.04 & 0.922 & 29.16 & 0.904 \\
    \midrule
    \textbf{(Ours)} & \textcolor{best}{26.16} & \textcolor{best}{0.928} & \textcolor{best}{29.81} & \textcolor{best}{0.908} \\
    \bottomrule
\end{tabular*}\vspace{-2mm}
\end{table}

Table~\ref{tab:guidance-ablation-review} shows that adding degradation descriptions improves PSNR by 0.53~dB on UIEB but only 0.03~dB on LSUI. CLIP fine-tuning offers no consistent advantage. TDD achieves the best PSNR and SSIM on both datasets, exceeding the strongest alternative by 0.90~dB on UIEB and 0.61~dB on LSUI.

\begin{figure}[!tb]
\centering
\includegraphics[width=\columnwidth]{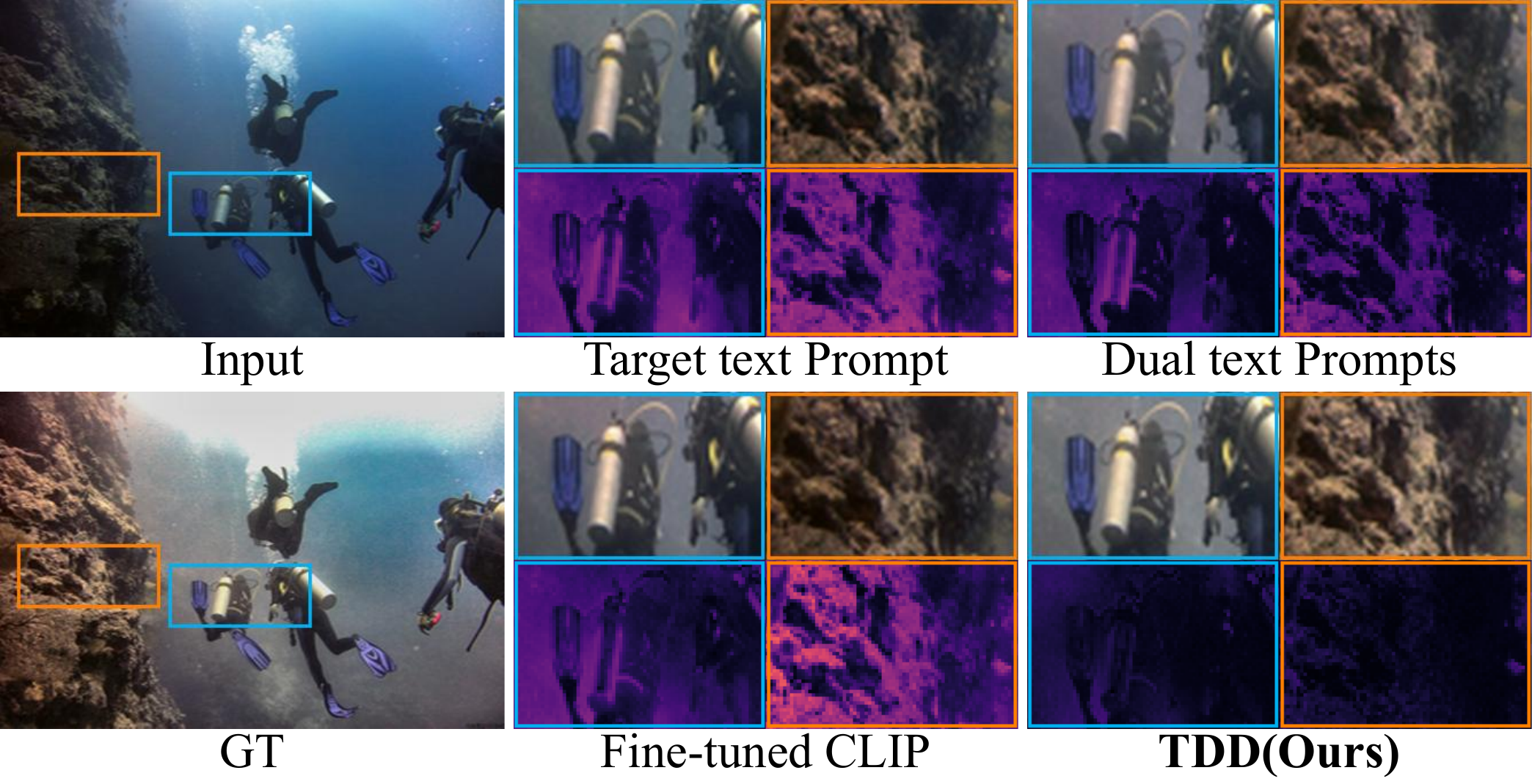}\vspace{-2mm}
\caption{Guidance comparison on UIEB: enhanced images, reference-relative MAE maps, and local enlargements.}\vspace{-2mm}
\label{fig:guidance-ablation}
\end{figure}

In Fig.~\ref{fig:guidance-ablation}, TDD restores warm rock-wall tones and blue--cyan water closer to the reference than the purplish text-based results, with lower MAE across the wall, divers, and water.

\begin{figure}[!tb]
\centering
\includegraphics[width=\columnwidth]{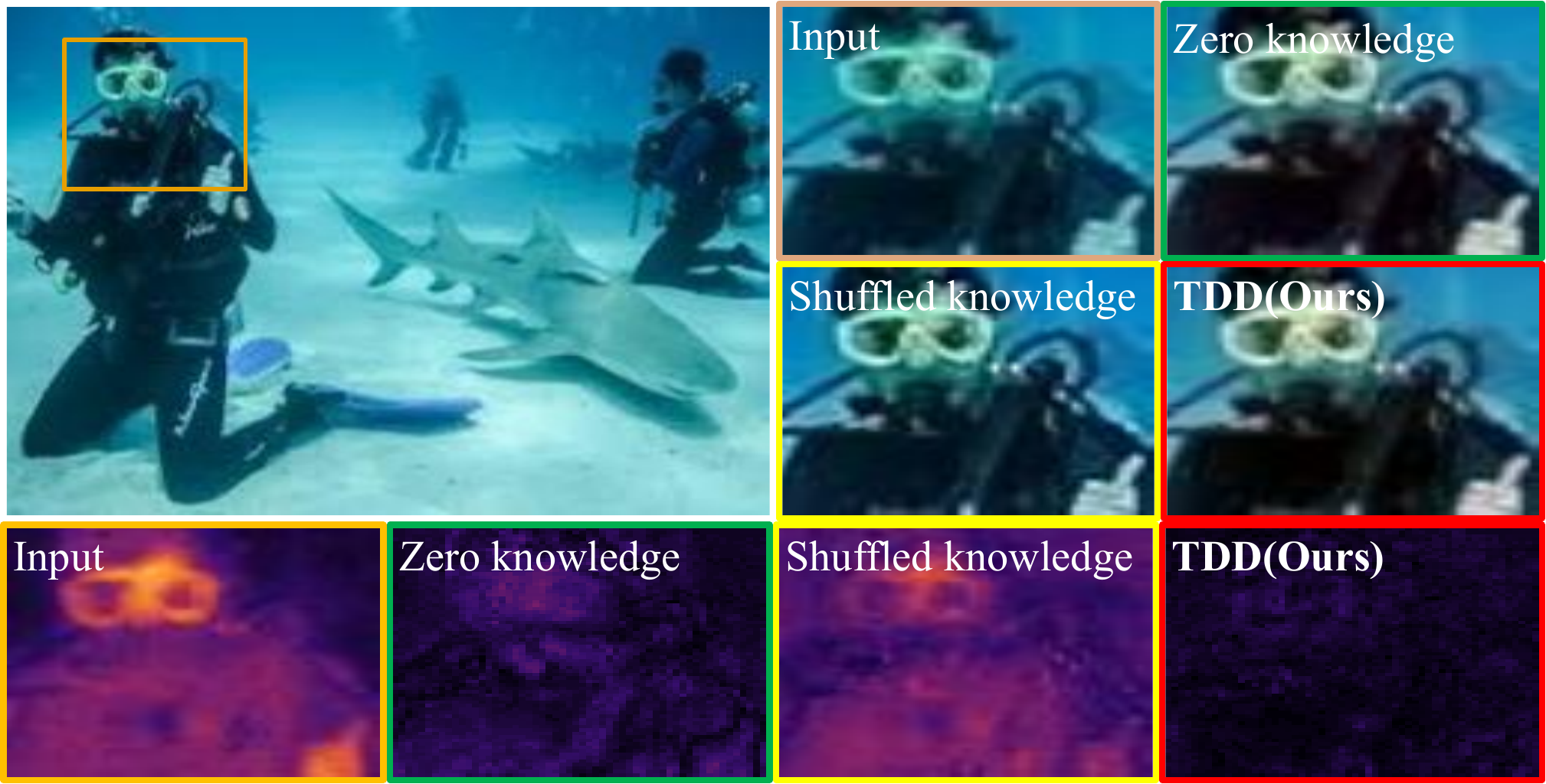}
\caption{Knowledge validity on UIEB with zero, shuffled, and input-matched knowledge. The bottom row shows local CIEDE2000 ($\Delta E_{00}$) color differences from the reference in CIELAB space~\cite{sharma2005ciede2000}; brighter colors indicate larger errors. Border colors identify configurations.}\vspace{-2mm}
\label{fig:knowledge-validity}
\end{figure}

\noindent \textbf{Contribution of Knowledge.}
With SKF, correctly matched distillation supervision, and all other training settings fixed, we retrain the enhancement network using zero or shuffled knowledge. \emph{Zero knowledge} sets both inputs to zero; \emph{Shuffled knowledge }jointly permutes T/D inputs so each image receives another image's paired knowledge. Both interventions remain active at evaluation, where each test image uses paired T/D knowledge from a fixed training image in the shuffled setting.

\begin{table}[!tb]
\centering
\captionsetup{skip=10pt}
\caption{Knowledge-conditioning ablation on UIEB and LSUI with SKF retained. Best in \textcolor{best}{red}.}\vspace{-2mm}
\label{tab:knowledge-validity}
\footnotesize
\setlength{\tabcolsep}{3pt}
\renewcommand{\arraystretch}{0.90}
\begin{tabular*}{0.87\columnwidth}{@{\extracolsep{\fill}}lcccc@{}}
    \toprule
    \multirow{2}{*}{Knowledge}
    & \multicolumn{2}{c}{UIEB}
    & \multicolumn{2}{c}{LSUI} \\
    \cmidrule(lr){2-3}\cmidrule(lr){4-5}
    & PSNR $\uparrow$ & SSIM $\uparrow$
    & PSNR $\uparrow$ & SSIM $\uparrow$ \\
    \midrule
    Zero knowledge & 25.18 & 0.920 & 28.98 & 0.898 \\
    Shuffled knowledge & 24.15 & 0.911 & 28.42 & 0.893 \\
    \midrule
    \textbf{TDD (Ours)}
    & \textcolor{best}{26.16} & \textcolor{best}{0.928}
    & \textcolor{best}{29.81} & \textcolor{best}{0.908} \\
    \bottomrule
\end{tabular*}\vspace{-2mm}
\end{table}

In Table~\ref{tab:knowledge-validity}, TDD improves PSNR over \emph{zero knowledge} by 0.98~dB on UIEB and 0.83~dB on LSUI. \emph{shuffled knowledge }performs worse than \emph{zero knowledge}, indicating that mismatched conditioning can interfere with restoration.

In Fig.~\ref{fig:knowledge-validity}, TDD reduces the cyan cast and color errors around the diver's face and mask compared with both zero and shuffled knowledge.

\begin{figure}[!tb]
\centering
\includegraphics[width=\columnwidth]{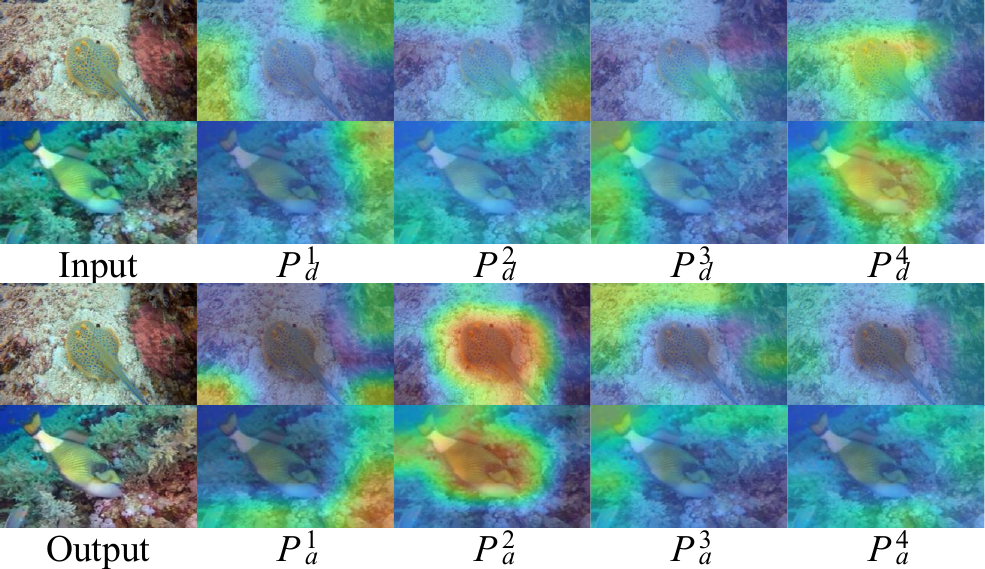}\vspace{-2mm}
\caption{Regional assignment responses on LSUI: discrepancy knowledge on inputs (top) and target-appearance knowledge on outputs (bottom). Numbers index regional vectors.}\vspace{-2mm}
\label{fig:knowledge-responses}
\end{figure}

Figure~\ref{fig:knowledge-responses} provides a complementary spatial view: the second appearance vector follows the fish's shape and orientation, while the fourth discrepancy vector responds over a different extent, with other vectors emphasizing surrounding regions.

\begin{table}[!tb]
\centering
\captionsetup{skip=10pt}
\caption{Knowledge distillation on UIEB. Parameters and FLOPs are reported for the knowledge generators. Best in \textcolor{best}{red}.}\vspace{-2mm}
\label{tab:distillation-ablation}
\footnotesize
\setlength{\tabcolsep}{2pt}
\renewcommand{\arraystretch}{0.95}
\begin{tabular}{@{}lcccc@{}}
    \toprule
    \multirow{2}{*}{Distillation}
    & \multicolumn{2}{c}{UIEB}
    & \multicolumn{2}{c}{Generator cost} \\
    \cmidrule(lr){2-3}\cmidrule(lr){4-5}
    & PSNR $\uparrow$ & SSIM $\uparrow$
    & Params (M) $\downarrow$ & FLOPs (G) $\downarrow$ \\
    \midrule
    Separate
    & 26.11 & 0.927 & 36.205 & 4.843 \\
    \textbf{Joint (Ours)}
    & \textcolor{best}{26.16}
    & \textcolor{best}{0.928}
    & \textcolor{best}{23.431}
    & \textcolor{best}{2.458} \\
    \bottomrule
\end{tabular}\vspace{-2mm}
\end{table}

\begin{figure}[!tb]
\centering
\includegraphics[width=\columnwidth]{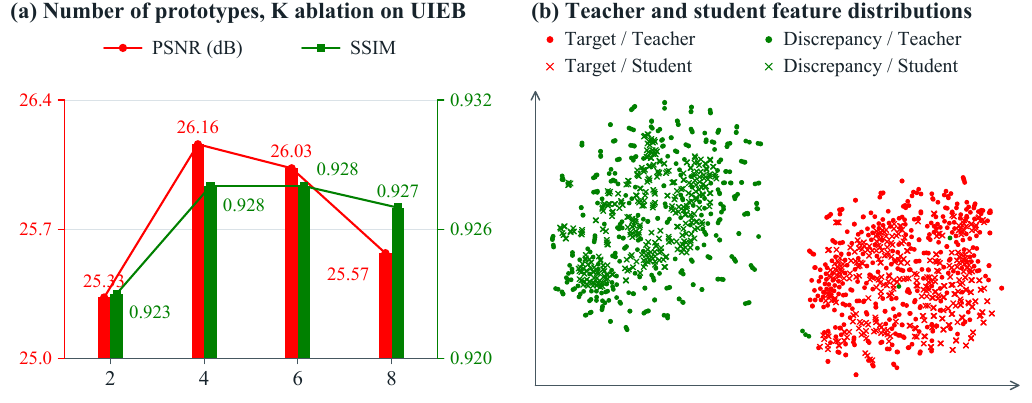}\vspace{-2mm}
\caption{Analysis of knowledge distillation.
(a) PSNR and SSIM on UIEB for different numbers of prototypes $K$.
(b) t-SNE visualization of teacher and student features for target-appearance and discrepancy knowledge.}\vspace{-2mm}
\label{fig:prototype-ablation-joint-roles}
\end{figure}

\noindent \textbf{Knowledge Distillation.}
Figure~\ref{fig:prototype-ablation-joint-roles}(b) shows that teacher and student features of the same type occupy nearby regions in the t-SNE projection, while target-appearance and discrepancy features remain distinct.

Table~\ref{tab:distillation-ablation} shows that a shared generator achieves comparable restoration with a slight numerical gain and reduces generator parameters and FLOPs by about 35\% and 49\%. Joint supervision from the same degraded--reference pair may encourage shared features useful for both predictions, potentially explaining the small gain.

Figure~\ref{fig:prototype-ablation-joint-roles}(a) compares $K\in\{2,4,6,8\}$ on UIEB. We use $K=4$ by default, as it gives the highest PSNR and ties with $K=6$ for the highest SSIM at the reported precision.

\noindent \textbf{Progressive Knowledge Fusion.}
\begin{table}[!tb]
\centering
\caption{SKF component ablation on UIEB. T/D denote target-appearance/discrepancy inputs to SKF. FLOPs are measured at $256\times256$. Best in \textcolor{best}{red}.}\vspace{-2mm}
\label{tab:skf-components}
\footnotesize
\setlength{\tabcolsep}{2pt}
\renewcommand{\arraystretch}{0.95}
\begin{tabular*}{0.85\columnwidth}{@{\extracolsep{\fill}}lcccc@{}}
    \toprule
    SKF configuration
    & PSNR $\uparrow$
    & SSIM $\uparrow$
    & Params (M)
    & FLOPs (G) \\
    \midrule
    Baseline (w/o SKF)
    & 24.61 & 0.913 & 15.55 & 177.34 \\
    D-only SKF
    & 25.73 & 0.923 & 41.12 & 185.23 \\
    T-only SKF
    & 25.62 & 0.922 & 40.30 & 180.69 \\
    \midrule
    \textbf{TDD (Ours)}
    & \textcolor{best}{26.16} & \textcolor{best}{0.928}
    & 41.86 & 185.74 \\
    \bottomrule
\end{tabular*}\vspace{-2mm}
\end{table}
In Table~\ref{tab:skf-components}, both single-branch variants (\emph{D-only} and \emph{T-only}) improve over the baseline, while combining them achieves the best PSNR and SSIM, exceeding \emph{D-only} and \emph{T-only} by 0.43 and 0.54~dB in PSNR, respectively.

\begin{figure}[!t]
\centering
\includegraphics[width=\columnwidth]{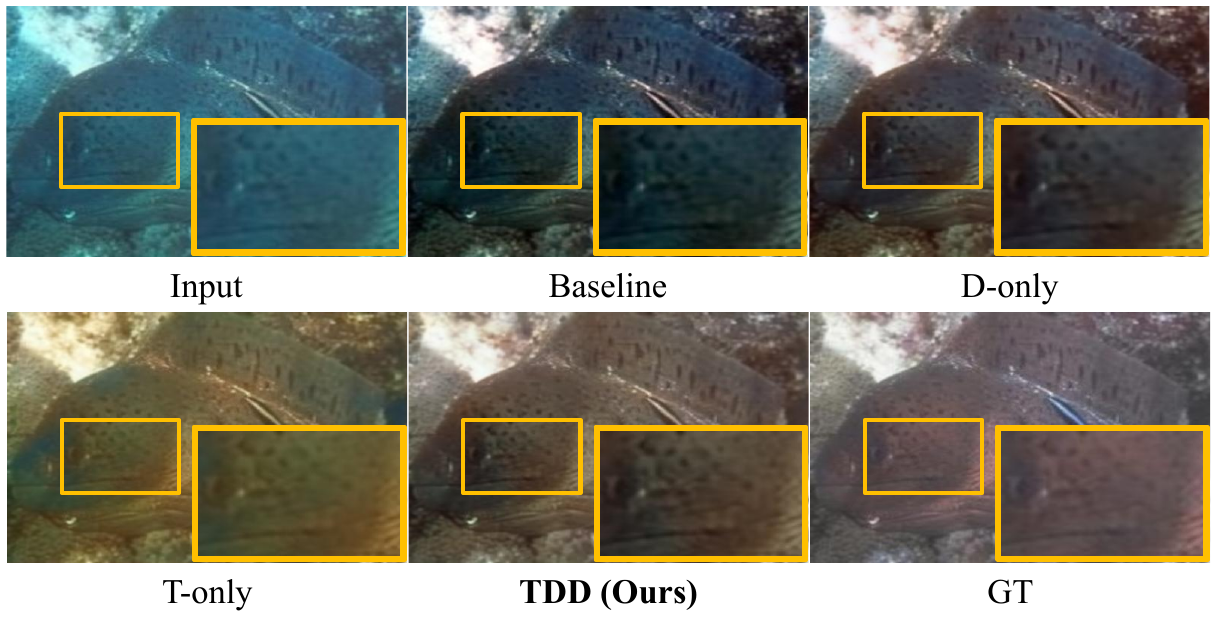}\vspace{-2mm}
\caption{Qualitative SKF component ablation on a scene with local enlargements. }\vspace{-2mm}
\label{fig:skf-components}
\end{figure}

Figure~\ref{fig:skf-components} shows that combining both branches reduces the baseline's blue-green cast and preserves fish markings, supporting discrepancy-conditioned correction followed by target-guided refinement.

In Table~\ref{tab:semuie-skf-uieb-best}, \emph{Concat} and \emph{Add} spatially broadcast student knowledge and fuse it with local image features via concatenation or addition, followed by convolutional fusion and the unchanged Restormer block. \emph{D $\rightarrow$ T} achieves the highest PSNR and SSIM. Reversing the order lowers PSNR by 0.84~dB at identical parameter and FLOP counts, supporting discrepancy-conditioned correction before target-guided refinement.

\begin{table}[!tb]
\centering
\caption{Fusion strategies ablation on UIEB. Best in \textcolor{best}{red}. }\vspace{-2mm}
\label{tab:semuie-skf-uieb-best}
\footnotesize
\setlength{\tabcolsep}{2pt}
\renewcommand{\arraystretch}{0.95}
\begin{tabular*}{\columnwidth}{@{\extracolsep{\fill}}lcccc@{}}
    \toprule
    Fusion strategy
    & PSNR $\uparrow$
    & SSIM $\uparrow$
    & Params (M)
    & FLOPs (G) \\
    \midrule
    Concat
    & 25.76 & 0.922 & 40.63 & 181.84 \\
    Add
    & 25.49 & 0.921 & 40.41 & 181.21 \\
    Joint fusion
    & 25.67 & 0.922 & 40.39 & 181.76 \\
    T $\rightarrow$ D
    & 25.32 & 0.921 & 41.86 & 185.74 \\
    \midrule
    \textbf{D $\boldsymbol{\rightarrow}$ T (TDD (Ours))}
    & \textcolor{best}{26.16} &\textcolor{best}{0.928}
    & 41.86 & 185.74 \\
    \bottomrule
\end{tabular*}\vspace{-2mm}
\end{table}

\begin{figure}[!tb]
\centering
\includegraphics[width=\columnwidth]{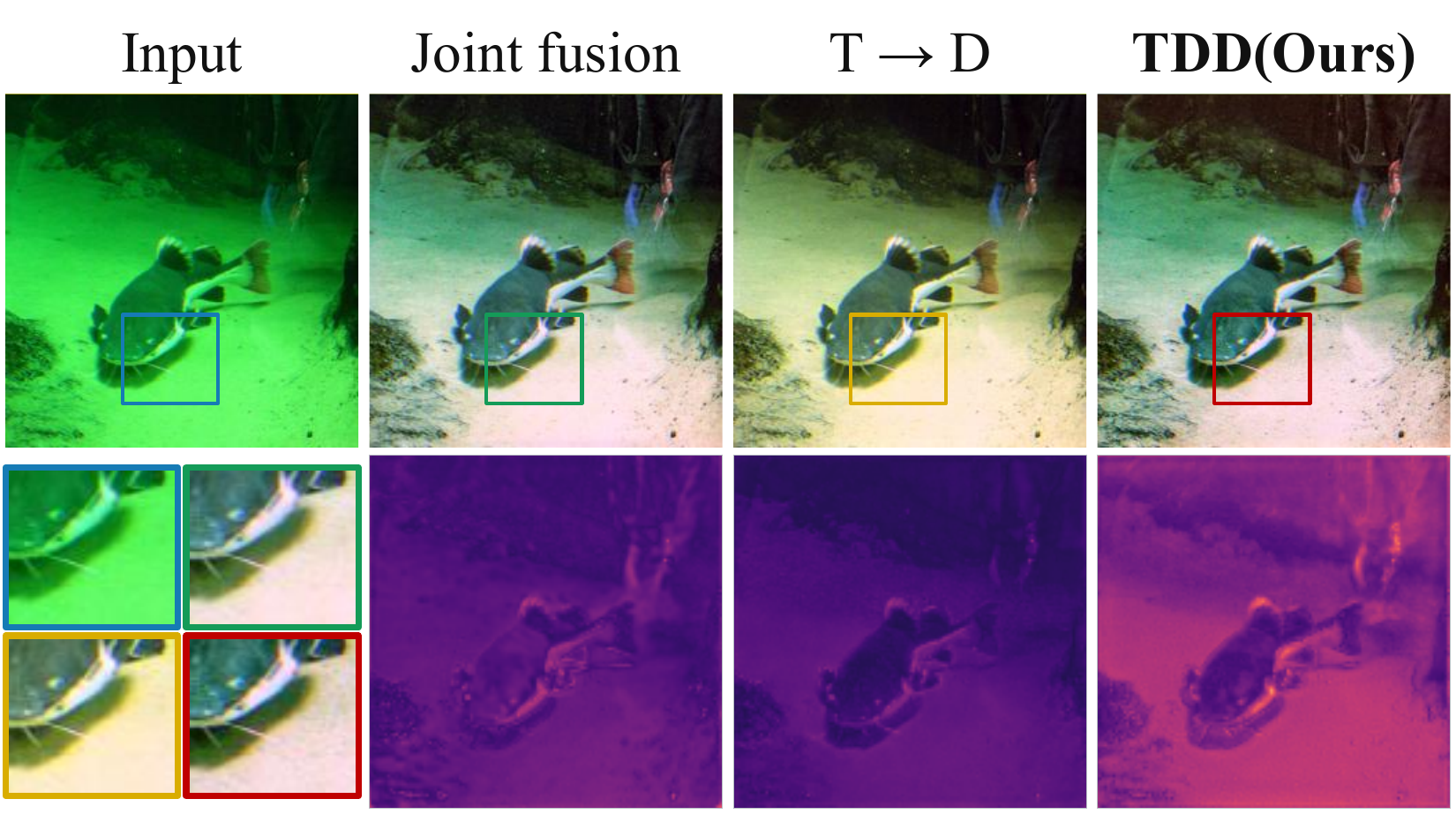}\vspace{-2mm}
\caption{Fusion comparison on C60 and U45: enhanced images, local enlargements, and corresponding feature maps.}\vspace{-2mm}
\label{fig:fusion-ablation}
\end{figure}

In Fig.~\ref{fig:fusion-ablation}, \emph{D $\rightarrow$ T} reduces yellow casts while preserving fish-head markings, mouth boundaries, and fine barbels. The reversed order leaves a yellow seabed tint in the second example. Feature responses follow fish contours and surrounding structures.
Additional visual comparisons and feature visualizations are provided in the supplementary material.
\section{Conclusion}
\label{sec:conclusion}

In this paper, we presented target--discrepancy knowledge distillation (TDD) for underwater image enhancement, motivated by the observation that underwater degradation can distort not only image appearance but also the semantic cues used to guide restoration. Rather than relying solely on descriptions of degraded scene content, TDD jointly learns two complementary forms of visual knowledge: target-appearance knowledge that specifies the desired appearance and discrepancy knowledge that characterizes how the degraded observation deviates from that target. By distilling both representations from paired images into a lightweight guidance network, TDD makes reference-based knowledge available through the degraded input alone, without requiring the teacher model or reference images at inference. Furthermore, the proposed progressive knowledge fusion translates their distinct supervisory roles into a correction-before-refinement process, in which discrepancy-guided spatial correction precedes target-guided appearance refinement. Extensive experiments on six underwater benchmarks demonstrate the effectiveness of the proposed framework across full-reference and no-reference evaluation metrics. Ablation studies and feature visualizations further validate the contributions of the knowledge representations, their joint distillation, and their progressive integration. 

\section{Limitations}
\label{sec:limitations}
Although reference images are unnecessary at inference, TDD relies on paired degraded–reference images to construct training supervision, as in many supervised UIE methods. Extending knowledge distillation to unpaired settings remains an important direction for future work.
\section*{LLM Usage Statement}
Generative AI tools were used to polish wording, correct grammar, improve formatting, and make sentences read more smoothly. The authors take responsibility for the final content.

\maketitlesupplementary
\setcounter{section}{0}
\setcounter{figure}{0}
\setcounter{table}{0}
\setcounter{equation}{0}
\renewcommand{\thesection}{S\arabic{section}}
\renewcommand{\thefigure}{S\arabic{figure}}
\renewcommand{\thetable}{S\arabic{table}}
\renewcommand{\theequation}{S\arabic{equation}}
\captionsetup[figure]{skip=5pt}

\section{Method Details}
\label{sec:supp-method-details}

\noindent\textbf{Compact Knowledge Construction.}
The frozen NAUTILUS teacher~\cite{xu2025nautilus} yields a global descriptor and a $4\times4$ regional grid per image. At grid position $j$, let $z_j^I,z_j^Y$ denote the $\ell_2$-normalized features of the degraded image $I$ and reference $Y$. We construct two complementary representations:
$u_j^{\mathrm a}=z_j^Y$ for target appearance and
$u_j^{\mathrm d}=z_j^I-z_j^Y$ for discrepancy.
A shared $W_0\in\mathbb R^{3584\times512}$ maps both to the student dimension:
$v_j^r=\mathcal N(u_j^rW_0)$, where
$r\in\{\mathrm a,\mathrm d\}$ and $\mathcal N$ denotes $\ell_2$ normalization.
We obtain $W_0$ from the orthonormal $Q$ factor of a Gaussian matrix using reduced QR decomposition (seed 20260825, used only for generating this fixed projection and unrelated to the training seed), and keep it fixed during training. The global descriptor undergoes the same projection and normalization but is not clustered.

We cluster the 16 regional features independently for each image and knowledge branch using weighted spherical $K$-means. Appearance features receive uniform weights. For discrepancy features, we use
$w_j^{\mathrm d}=\delta_j/\max(\bar\delta,10^{-6})$,
where $\delta_j=\|z_j^I-z_j^Y\|_2$ and $\bar\delta$ is the mean deviation over the 16 regions. Measuring the deviation before projection and normalization preserves its magnitude. We floor the weights at $10^{-4}$. The first center is initialized with the region farthest from the weighted mean; subsequent centers are initialized by selecting the region farthest from its nearest existing center, using the weighted cosine distance
$w_j^r[1-\cos(v_j^r,c)]$.
We perform eight iterations, assigning regions by maximum cosine similarity and updating each center as
\begin{equation}
  p_k^r=\mathcal N\!\left(
  \sum_{j\in\Omega_k^r}w_j^r v_j^r
  \right),
  \label{eq:supp-centers}
\end{equation}
where $\Omega_k^r$ denotes the regions assigned to the $k$-th cluster. Empty clusters are reseeded using the same farthest-region rule. With the default $K=4$, each branch contains one global descriptor and four regional prototypes, all 512-dimensional.

\smallskip\noindent\textbf{Prototype Matching and Distillation.}
Prototype indices do not represent fixed semantic categories across images. Therefore, within each image and branch, we match the student predictions $\widehat p_k^r$ to teacher prototypes through a one-to-one assignment:
\begin{equation}
  \sigma^\star=\arg\max_{\sigma\in\mathfrak S_K}
  \sum_{k=1}^{K}
  \cos(\widehat p_k^r,p_{\sigma(k)}^r),
  \label{eq:supp-matching}
\end{equation}
where $\mathfrak S_K$ contains all permutations (24 at $K=4$). Global descriptors are aligned directly, while regional prototypes are compared after this matching.

For each knowledge branch $r\in\{\mathrm d,\mathrm a\}$, the distillation objective combines the global descriptor distance and the mean distance over matched regional prototypes:
\begin{equation}
\begin{aligned}
  \mathcal L_r
  ={}& 1-\cos(\widehat p_g^r,p_g^r)\\
  &+\frac{0.8}{K}\sum_{k=1}^{K}
  \left[1-\cos(
  \widehat p_k^r,p_{\sigma^\star(k)}^r
  )\right].
\end{aligned}
\label{eq:supp-distill}
\end{equation}
All matched regional prototypes contribute equally. The two branch objectives are combined as
\begin{equation}
  \mathcal L_{\mathrm{distill}}
  =\mathcal L_d+\mathcal L_a.
  \label{eq:supp-total-distill}
\end{equation}
The target-appearance objective transfers reference-derived appearance representations, whereas the discrepancy objective teaches the student to estimate the corresponding observation-to-target feature deviations.

In parallel with prototype-vector prediction, the knowledge generator predicts a weight for each of the $K$ regional prototypes in every branch. Specifically, each decoded prototype latent vector is fed into two parallel linear heads: one predicts the prototype embedding, followed by $\ell_2$ normalization, and the other predicts a scalar weight logit. A branch-wise softmax over the $K$ logits produces the prototype weights $\{\widehat w_k^r\}_{k=1}^{K}$, which sum to one. These weights provide an image-level prior that guides spatial knowledge association.

\smallskip\noindent\textbf{Spatial Knowledge Association.}
In Eq.~(6) of the main paper, let $e_s(i,j)$ denote the association score between image location $i$ and knowledge vector $j$ in the selected representation $\widehat P_s$, i.e., the pointwise value of the similarity function $\mathcal S_s$. Let $q_{si}$ and $k_{sj}$ denote the projected image query and knowledge key, respectively, and let $V_s=\widehat P_sW_V^s$ contain the projected knowledge values. The association scores and spatial guidance are computed as
\begin{align}
  e_s(i,j)
  &=\tau_s\cos(q_{si},k_{sj})
    +\beta_s\log\pi_{sj},
  \label{eq:supp-similarity}\\
  A_s(i,:)&=\operatorname{softmax}(e_s(i,:)),
  \qquad H_s=A_sV_s,
  \label{eq:supp-readout}
\end{align}
where $\tau_s>0$ is a learnable similarity scale and $\beta_s\geq0$ controls the contribution of the predicted knowledge weights. The prior assigns $0.35$ to the global descriptor and distributes the remaining $0.65$ among the regional prototypes in proportion to their predicted weights:
\begin{equation}
  \pi_{s,j}=
  \begin{cases}
    0.35, & j=\mathrm{global},\\
    0.65\,\widehat w_j^r, & j\in\mathrm{regional},
  \end{cases}
  \label{eq:supp-weight-prior}
\end{equation}
where $r=\mathrm d$ for the discrepancy step and $r=\mathrm a$ for the target-appearance step. The predicted weights provide an image-level bias shared across spatial locations, while the cosine similarity captures the location-specific correspondence. Softmax is applied over the $K+1$ knowledge vectors at each location, so the resulting attention remains dependent on the current image features.

\smallskip\noindent\textbf{Degradation Strength Estimation.}
Besides selecting relevant knowledge, SKF estimates the update strength required at each spatial location. We combine a local feature score with a global degradation-strength prediction from the knowledge generator. Given the incoming feature map at the current scale, a local prediction branch applies a $3\times3$ depth-wise convolution, GELU activation, and a $1\times1$ convolution to produce a scalar score $\ell_i$ at each location. This branch and the local feature extraction branch receive the same incoming feature map; they provide separate spatial-strength and local-feature information.

Meanwhile, the knowledge generator predicts a non-negative global strength $\rho$ from the input image. We transform it into a bounded image-level value:
\begin{equation}
  p=\operatorname{clip}(1-e^{-\rho},0.001,0.999).
  \label{eq:supp-global-strength}
\end{equation}
The location-dependent degradation strength is then
\begin{equation}
  d_i=\sigma\left(
  \ell_i+0.35\,\operatorname{logit}(p)
  \right),
  \label{eq:supp-local-strength}
\end{equation}
where $\sigma$ denotes the sigmoid function. The local score determines the spatial variation in degradation strength, whereas the global prediction provides a shared image-level bias. Importantly, $d_i$ is estimated from image features and the global prediction, independently of the knowledge attention map and reference images. It controls update strength rather than knowledge selection.

\smallskip\noindent\textbf{Spatial Feature Update.}
Let $B_s$ denote the incoming backbone feature map and $U_s$ the fused feature at the output of SKF. Let $\Delta_s(U_s)$ denote the branch update after refinement and a $1\times1$ projection back to the channel dimension of $B_s$. The update is added to the incoming features as
\begin{equation}
  B_{s+1}=B_s+\eta_s\mathcal I_s\!\left[
    g_s\odot\Delta_s(U_s)\right],
  \label{eq:supp-skf-update}
\end{equation}
where $\eta_s$ is a learned, tanh-bounded scalar and $\mathcal I_s$ bilinearly resizes the update from the association grid (at most $64\times64$) when necessary. The spatial weight $g_s$ is shared across channels. For discrepancy-conditioned correction, the gate combines the estimated degradation strength with the strongest knowledge association:
\begin{equation}
  g_1(i)=d_i\left[
  0.35+0.65\max_j A_1(i,j)
  \right].
  \label{eq:supp-discrepancy-gate}
\end{equation}
For target-appearance refinement, the same degradation estimate is reused:
\begin{equation}
  g_2(i)=0.35+0.65d_i.
  \label{eq:supp-appearance-gate}
\end{equation}
Thus, knowledge attention determines which prototype is retrieved, while $d_i$ controls the strength of the spatial update. Their combination enables discrepancy-conditioned correction followed by target-guided refinement. All coefficients ($0.35/0.65$, $0.8$, $0.35$) are fixed empirically across experiments.

\smallskip\noindent\textbf{Input-dependent Knowledge in Stage 2.}
After Stage 1, the knowledge generator $\mathcal G_{\theta^\star}$ retains its learned parameters and remains in evaluation mode, outside the optimizer. Its predictions $(\widehat P^{\mathrm d},\widehat P^{\mathrm a})$, including the associated prototype-weight and global-strength predictions, are recomputed for each degraded input $I$ and can change as the visual encoder $E_\phi$ adapts. Gradients pass through the fixed generator to the visual encoder; only the encoder and restoration backbone are updated.

\section{Additional Visual Results}
\subsection{Knowledge and Feature Responses}
\begin{figure}[!tb]
  \centering
  \includegraphics[width=\linewidth]{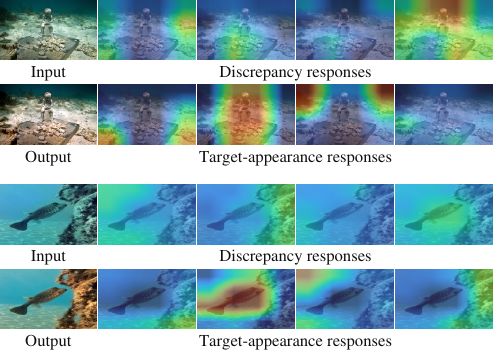}
  \caption{Regional responses on UIEB~\cite{li2019underwater} and EUVP~\cite{9001231}. The input and output rows show the four discrepancy and four target-appearance responses, respectively. All maps are overlaid on the same input pixels using a fixed probability scale.}
  \label{fig:regional}
\end{figure}

Figure~\ref{fig:regional} shows that target-appearance and discrepancy prototypes attend to different structures even on the same input. In the statue scene, an appearance prototype localizes on the statue and its base, whereas a discrepancy prototype responds to the adjacent seabed where the observation deviates from the reference. In the fish scene, an appearance response traces the body from tail to head, while a discrepancy response extends along the surrounding rock face. This separation matches the intended division of roles: appearance vectors encode where the restored content should be, and discrepancy vectors encode where and how the input must be corrected. The spatial responses corroborate the component ablation and explain why keeping the two branches separate yields complementary guidance.

\begin{figure}[!tb]
  \centering
  \includegraphics[width=\linewidth]{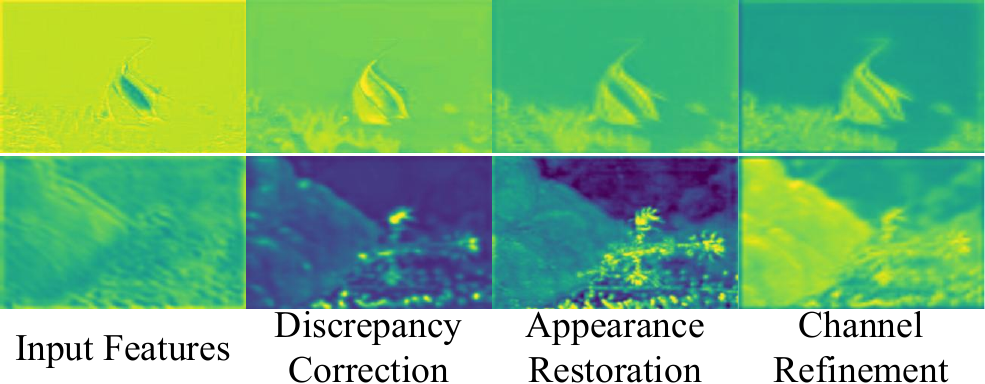}
  \caption{Feature responses of the same channel across successive PKF stages: input features, discrepancy-conditioned correction, target-appearance refinement, and channel refinement. The discrepancy step activates localized regions, the target step recovers finer structures, and channel refinement consolidates both.}
  \label{fig:feature}
\end{figure}

Figure~\ref{fig:feature} further shows how the two SKF steps reshape backbone features. The discrepancy-conditioned response highlights a few localized regions, concentrating on the areas that require correction, whereas the appearance-conditioned response brings out finer structures across the right side. After channel refinement, these local responses are consolidated with broader structural information, including the diagonal boundary on the left. The two updates thus supply different feature cues---one local and corrective, the other structural and restorative---which are subsequently integrated through channel interactions. This division of labor motivates the correction-before-refinement ordering in SKF.

\begin{figure}[!tb]
  \centering
  \includegraphics[width=\linewidth]{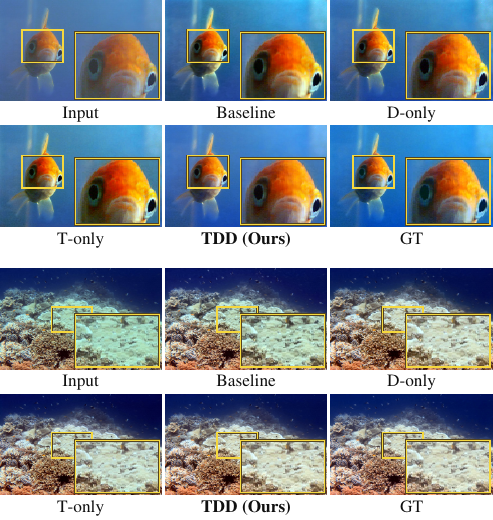}
  \caption{SKF comparisons on UIEB~\cite{li2019underwater} with matched local enlargements. D-only/T-only denote discrepancy-only/target-appearance-only SKF; the full configuration removes the residual color cast while preserving foreground detail.}
  \label{fig:skf}
\end{figure}

\subsection{Ablation Studies}
\noindent\textbf{Complementary Knowledge Branches.}
Figure~\ref{fig:skf} shows that discrepancy-only SKF removes much of the green tint retained by the baseline on the pale seabed, while combining both branches brings the sand closer to the reference. In the fish crop, target-only SKF leaves yellow coloration around the mouth, whereas TDD better preserves both the orange forehead and the paler mouth seen in the reference. The two branches act in sequence: discrepancy knowledge corrects observation-dependent deviations such as the green cast, while target-appearance knowledge restores scene-specific colors such as the fish markings. Consistent with these visual observations, combining the branches improves PSNR by $0.43$~dB over D-only SKF and $0.54$~dB over T-only SKF on UIEB.

\begin{figure*}[!t]
  \centering
  \includegraphics[width=\textwidth]{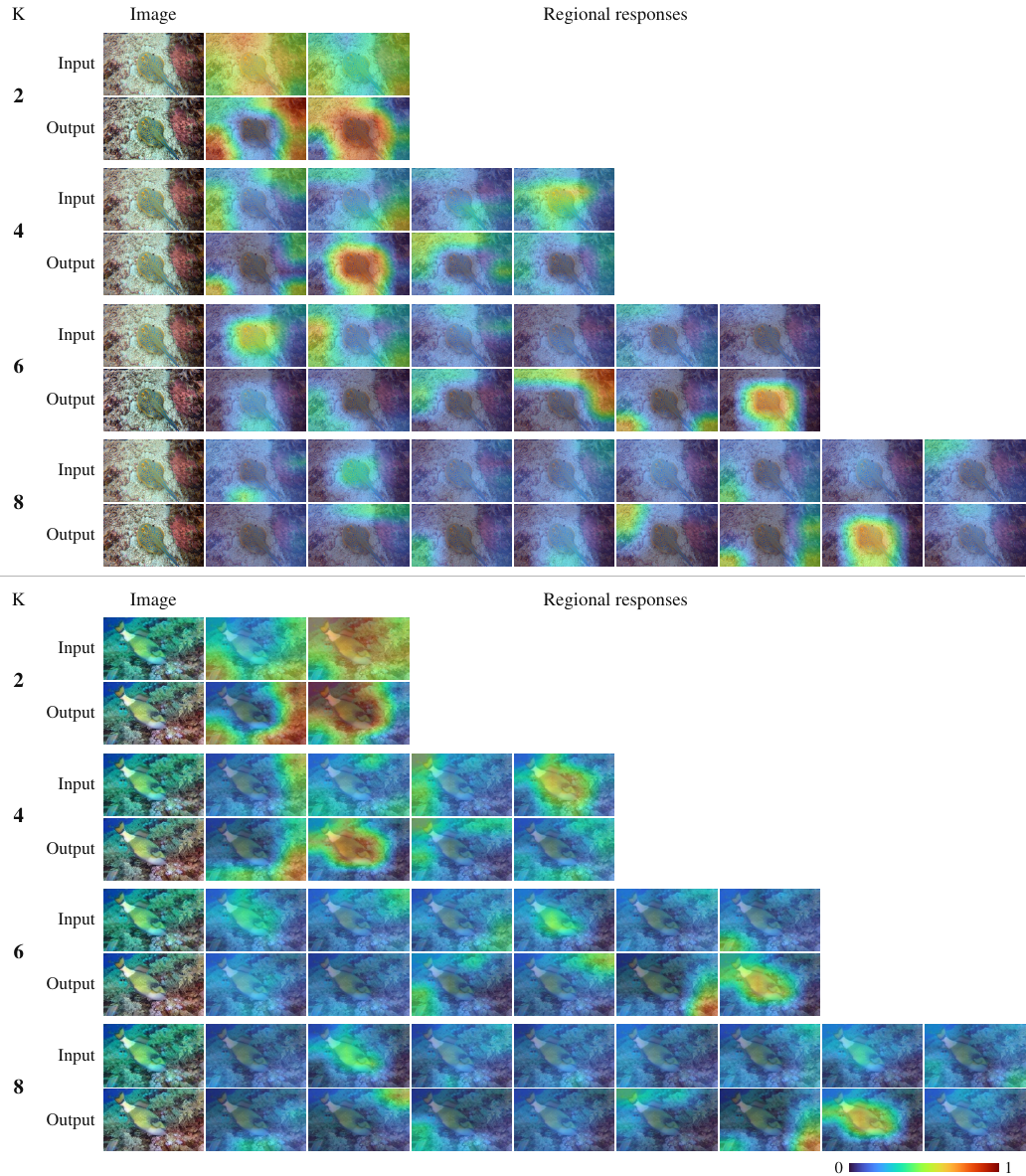}
  \caption{Prototype counts ($K=2,4,6,8$) on UIEB~\cite{li2019underwater} and LSUI~\cite{peng2023u}. The input and output rows show discrepancy and target-appearance responses, respectively, both overlaid on the input. The native $4\times4$ maps are normalized over the $K$ regional vectors and enlarged by bilinear interpolation. The $[0,1]$ color scale is fixed across examples, without per-image contrast normalization.}
  \label{fig:k-visual}
\end{figure*}

\noindent\textbf{Regional Prototype Count.}
Figure~\ref{fig:k-visual} shows how the number of regional prototypes affects spatial decomposition. With $K=2$, individual responses cover relatively broad regions that mix foreground and background. Increasing $K$ to 4 produces more differentiated assignments: an appearance response concentrates on the fish, while other prototypes emphasize surrounding reef regions. At $K=6$ and $K=8$, a similar fish-centered response remains, but the additional prototypes do not reveal consistently finer structure. Because responses are normalized over $K$, this comparison concerns their spatial distributions. The visual trend matches the quantitative ablation in the main paper, where $K=4$ attains the highest PSNR and ties $K=6$ on SSIM at the reported precision, supporting our default choice of $K=4$.

\begin{figure*}[!t]
  \centering
  \includegraphics[width=\columnwidth]{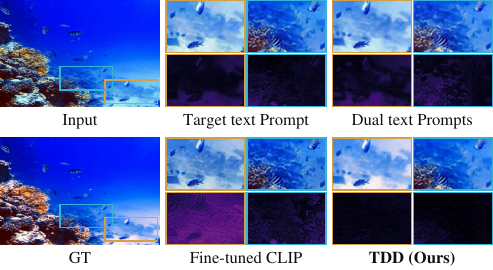}\hfill
  \includegraphics[width=\columnwidth]{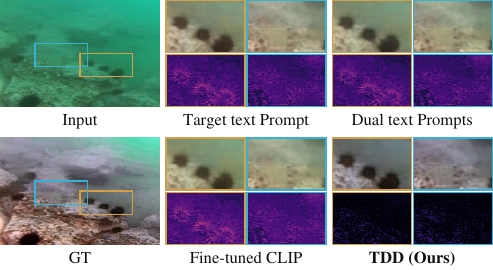}
  \caption{Guidance strategies on UIEB~\cite{li2019underwater} and LSUI~\cite{peng2023u}. Input and GT are on the left of each group. For each single-pass output, the upper pair shows two matched crops and the lower pair their RGB mean absolute errors relative to GT, on a common $[0,0.4]$ scale. Border colors identify corresponding crops.}
  \label{fig:guidance}
\end{figure*}

\begin{figure*}[!t]
  \centering
  \includegraphics[width=\textwidth]{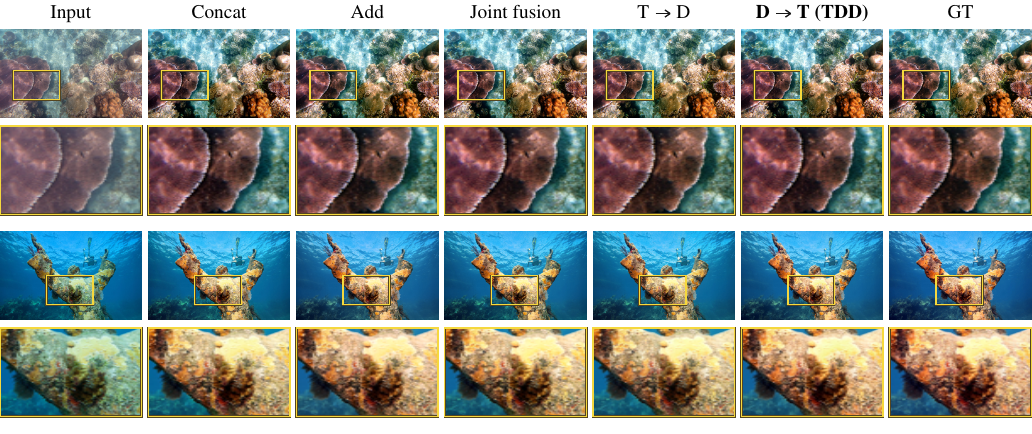}
  \caption{Fusion strategies on UIEB~\cite{li2019underwater} with identical crop coordinates: Input, Concat, Add, Joint fusion, $T\!\rightarrow\!D$, $D\!\rightarrow\!T$ (TDD), and the reference. T and D denote target-appearance and discrepancy knowledge; all strategies use the same parameters and FLOPs.}
  \label{fig:fusion}
\end{figure*}

\noindent\textbf{Guidance Representations.}
Figure~\ref{fig:guidance} compares different guidance strategies under the same enhancement architecture. In the first example, text-guided outputs retain a blue cast on the sand that appears pale in the reference, whereas TDD recovers the pale sand while preserving the blue--cyan water. The distinction is not simply whether blue is present, but where the corresponding appearance belongs. In the green seabed scene, prompt-based outputs retain yellow-green coloration around the dark regions, while TDD produces a grayer substrate. The matched error maps provide complementary evidence: in the gold-bordered LSUI crop, the mean absolute RGB error on $[0,1]$ is $0.016$ for TDD versus $0.089$ for dual text prompts. By representing the target appearance and its observation-dependent discrepancy explicitly, TDD assigns colors to the correct structures, which explains its markedly lower error.

\noindent\textbf{Fusion Strategies.}
Figure~\ref{fig:fusion} compares five fusion strategies under identical crop coordinates. \emph{Concat} and \emph{Add} broadcast student knowledge spatially and combine it with local image features through concatenation or addition, followed by convolutional fusion and the unchanged restoration block; \emph{Joint fusion} merges the two knowledge streams before injection. These broadcast variants retain a yellowish cast on the statue and soften the coral-plate edge. The sequential strategies better preserve the reference appearance, and $D\!\rightarrow\!T$ most faithfully recovers the reddish-brown mottling of the statue while keeping the coral edge intact. Because all strategies share the same parameter and FLOP budgets, the comparison isolates how knowledge is combined: applying target-appearance guidance to features that have already undergone discrepancy-conditioned correction gives the subsequent refinement a cleaner feature basis and improves UIEB PSNR by $0.84$~dB over the reverse order. The visual and quantitative results confirm the proposed correction-before-refinement sequence.

\subsection{Comparisons with Other Methods}
\begin{figure*}[!t]
  \centering
  \includegraphics[width=\textwidth]{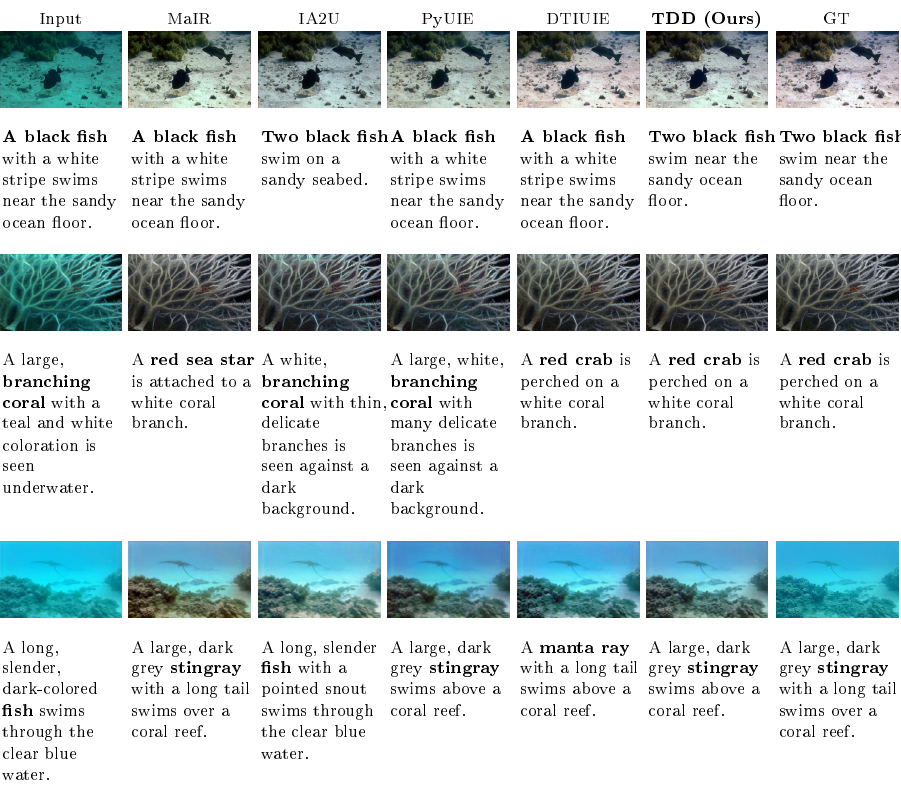}
  \caption{Language descriptions on UIEB~\cite{li2019underwater} and EUVP~\cite{9001231} for the input, four representative baselines, TDD, and the reference. The same frozen VLM receives each full image and the query: ``Describe the main subject and its observed colors in one sentence.'' Bold text marks count or category expressions in the original responses.}
  \label{fig:caption-selected}
\end{figure*}

\noindent\textbf{Semantic Cues.}
Figure~\ref{fig:caption-selected} compares descriptions generated by the same frozen captioner using identical prompts across seven columns: the input, four representative baselines, TDD, and the paired reference. The baselines cover distinct enhancement strategies: MaIR~\cite{li2025mair} employs a state-space model, IA2U~\cite{11267029} transfers semantic priors, PyUIE~\cite{11330175} uses a coarse-to-fine pyramid network, and DTIUIE~\cite{11433506} emphasizes perception-oriented restoration. The examples show differences beyond color vocabulary: the first scene highlights object-count consistency, the coral scene illustrates category recognition, and the open-water scene shows a change from a generic fish description to a stingray description. Descriptions of TDD outputs are more consistent with those of the references than descriptions generated from degraded inputs, showing that the restored appearance preserves object-level semantics obscured in the degraded input. These descriptions complement the image-level comparisons by confirming that the knowledge guiding pixel restoration also improves semantic interpretation.

\begin{figure*}[!t]
  \centering
  \includegraphics[width=\textwidth]{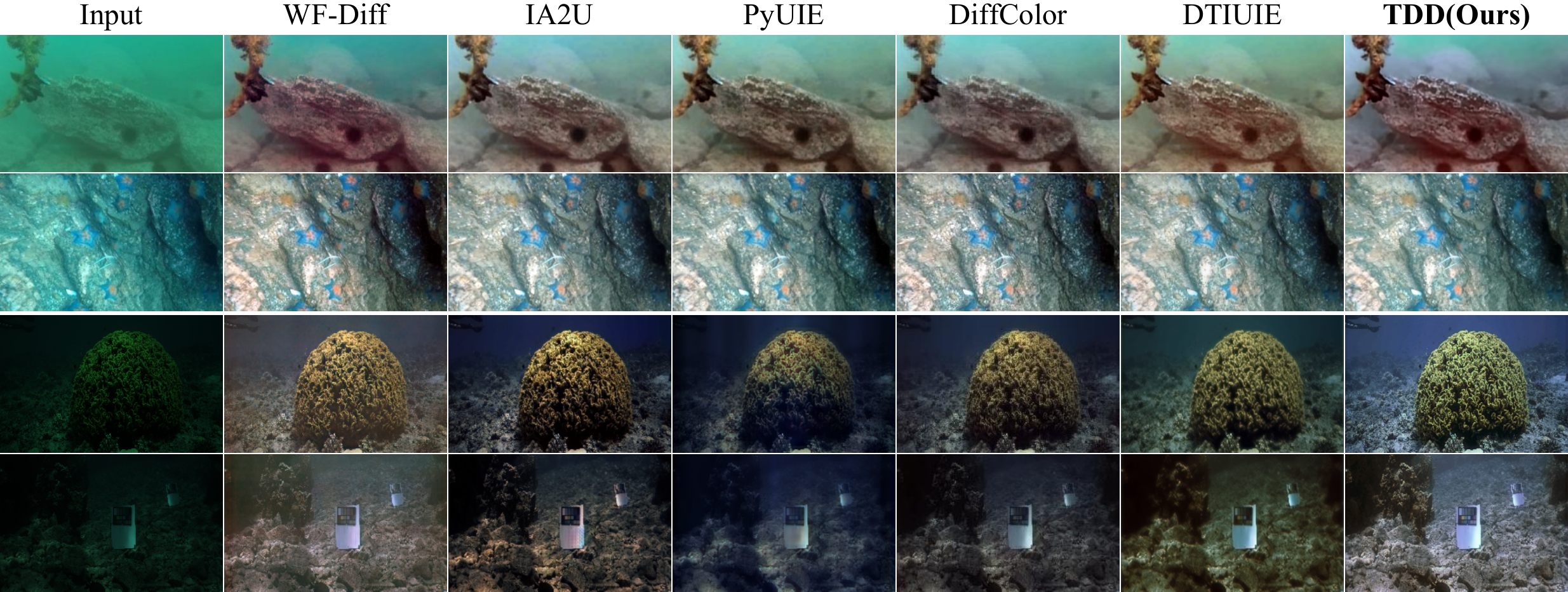}
  \caption{No-reference comparisons beyond C60: UCCS~\cite{8949763} (top two rows) and Sea-thru~\cite{8954437} (bottom two rows). TDD preserves local contrast and separates foreground from water without paired targets.}
  \label{fig:unpaired}
\end{figure*}

\noindent\textbf{No-reference Comparison.}
Figure~\ref{fig:unpaired} shows the input, five baselines, and TDD on UCCS~\cite{8949763} and Sea-thru~\cite{8954437} scenes beyond the paired evaluation examples, without a paired reference column. The five baselines are WF-Diff~\cite{zhao2024wavelet}, IA2U~\cite{11267029}, PyUIE~\cite{11330175}, DiffColor~\cite{11353942}, and DTIUIE~\cite{11433506}, covering frequency-domain diffusion, semantic prior transfer, coarse-to-fine pyramids, color restoration, and perception-oriented restoration. In the first UCCS scene, TDD better separates the reddish-brown foreground ridge from the cooler surrounding water, reducing the shared green cast. In the next scene, the blue sea star remains visually distinct as the cyan coloration of the surrounding stone recedes. In the darker Sea-thru scenes, local ridges near the base of the rounded coral are more distinguishable in TDD than in the shadowed PyUIE~\cite{11330175} output, while the light face of the color chart remains distinct against the textured seabed. These results show that the correction-before-refinement design generalizes to scenes without paired targets: TDD infers both the deviation to correct and the appearance to restore from the degraded input alone, producing clearer local contrast and color separation than prior methods.

{
    \small
    \bibliographystyle{ieeenat_fullname}
    \bibliography{ref}
}
\end{document}